# Policy Recognition in the Abstract Hidden Markov Model

**Hung H. Bui** BUIHH@CS.CURTIN.EDU.AU
**Svetha Venkatesh** SVETHA@CS.CURTIN.EDU.AU
**Geoff West** GEOFF@CS.CURTIN.EDU.AU
*Department of Computer Science*
*Curtin University of Technology*
*PO Box U1987, Perth, WA 6001, Australia*

## Abstract

In this paper, we present a method for recognising an agent's behaviour in dynamic, noisy, uncertain domains, and across multiple levels of abstraction. We term this problem *on-line plan recognition under uncertainty* and view it generally as probabilistic inference on the stochastic process representing the execution of the agent's plan. Our contributions in this paper are twofold. In terms of probabilistic inference, we introduce the Abstract Hidden Markov Model (AHMM), a novel type of stochastic processes, provide its dynamic Bayesian network (DBN) structure and analyse the properties of this network. We then describe an application of the Rao-Blackwellised Particle Filter to the AHMM which allows us to construct an efficient, hybrid inference method for this model. In terms of plan recognition, we propose a novel plan recognition framework based on the AHMM as the plan execution model. The Rao-Blackwellised hybrid inference for AHMM can take advantage of the independence properties inherent in a model of plan execution, leading to an algorithm for online probabilistic plan recognition that scales well with the number of levels in the plan hierarchy. This illustrates that while stochastic models for plan execution can be complex, they exhibit special structures which, if exploited, can lead to efficient plan recognition algorithms. We demonstrate the usefulness of the AHMM framework via a behaviour recognition system in a complex spatial environment using distributed video surveillance data.

## 1. Introduction

Plan recognition is the problem of inferring an actor's plan by watching the actor's actions and their effects. Often, the actor's behaviour follows a hierarchical plan structure. Thus, in plan recognition, the observer needs to infer about the actor's plans and sub-plans at different levels of abstraction in its plan hierarchy. The problem is complicated by the two sources of uncertainty inherent in the actor's planning process: (1) the stochastic aspect of plan refinement (a plan can be non-deterministically refined into different sub-plans), and (2) the stochastic outcomes of actions (the same action can non-deterministically result in different outcomes). Furthermore, the observer has to deal with a third source of uncertainty arising from the noise and inaccuracy in its own observation about the actor's plan. In addition, we would like our observer to be able to perform the plan recognition task "on-line" while the observations about the actor's plan are streaming in. We refer to this general problem as *on-line plan recognition under uncertainty.*

The seminal work in plan recognition (Kautz & Allen, 1986) considers a plan hierarchy, but does not deal with the uncertainty aspects of the problem. As a result, the approach can only postulate a set of possible plans for the actor, but is unable to determine which plan is more probable. Since then, the important role of uncertainty reasoning in plan recognition has been recognised (Charniak & Goldman, 1993; Bauer, 1994; van Beek, 1996), and Bayesian probability has been argued as the appropriate model (Charniak & Goldman, 1993; van Beek, 1996). The dynamic, "on-line" aspect of plan recognition has only been recently considered (Pynadath & Wellman, 1995, 2000; Goldman, Geib, & Miller, 1999; Huber, Durfee, & Wellman, 1994; Albrecht, Zukerman, & Nicholson, 1998). All of this recent work shares the view that online plan recognition is largely a problem of probabilistic inference in a stochastic process that models the execution of the actor's plan. While this view offers a general and coherent framework for modelling different sources of uncertainty, the stochastic process that we need to deal with can become quite complex, especially if we consider a large plan hierarchy. Thus, the main issue here is the computational complexity for dealing with this type of stochastic processes, and whether the complexity is scalable to more complex plan hierarchies.

### 1.1 Aim and Significance

In this paper, we demonstrate that the type of plan recognition problems described above scales reasonably well with respect to the number of levels of abstraction in the plan hierarchy. This is in contrast to the common-sense analysis that more levels in the plan hierarchy would introduce more variables to the stochastic process, which in turn, results in exponential complexity w.r.t the number of levels in the hierarchy.

In order to achieve this, we first assume a general stochastic model of plan execution that can model the three sources of uncertainty involved. The model for planning with a hierarchy of abstraction under uncertainty has been developed recently by the abstract probabilistic planning community (Sutton, Precup, & Singh, 1999; Parr & Russell, 1997; Forestier & Varaiya, 1978; Hauskrecht, Meuleau, Kaelbling, Dean, & Boutilier, 1998; Dean & Lin, 1995). To our advantage, we adopt their basic model, known as the *abstract Markov policies (AMP)* [1] as our model for plan execution. The AMP is an extension of a policy in Markov Decision Processes (MDP) that enables an abstract policy to invoke other more refined policies and so on down the policy hierarchy. Thus, the AMP is similar to a contingent plan that prescribes which sub-plan should be invoked at each applicable state of the world to achieve its intended goal, except that it can represent both the uncertainty in the plan refinement and in the outcomes of actions. Since an AMP can be described simply in terms of a state space and a Markov policy that selects among a set of other AMP's, using the AMP as the model for plan execution also helps us focus on the structure of the policy hierarchy.

The execution of an AMP leads to a special stochastic process which we called the *Abstract Markov Model* (AMM). The noisy observation about the environment state (e.g., the effects of action) can then be modelled by making the state "hidden", similar to the hidden state in the Hidden Markov Models (Rabiner, 1989). The result is an interesting and novel stochastic process which we term the *Abstract Hidden Markov Model*. Intuitively, the

1. Also known as *options*, *policies of Abstract Markov Decision Processes*, or *supervisor's policies*.

AHMM models how an AMP causes the adoption of other policies and actions at different levels of abstraction, which in turn generate a sequence of states and observations. In the plan recognition task, an observer is given an AHMM corresponding to the actor's plan hierarchy, and is asked to infer about the current policy being executed by the actor at all levels of the hierarchy, taking into account the sequence of observations currently available. This amounts to reversing the direction of causality in the AHMM, i.e. to determine a set of policies that can explain the sequence of observations at hand. We shall refer to this problem as *policy recognition*.

Viewing the AHMM as a type of dynamic Bayesian network (Dean & Kanazawa, 1989; Nicholson & Brady, 1992), it is known that the complexity of this kind of inferencing in the DBN depends on the size of the representation of the so-called *belief state*, the conditional joint distribution of the variables in the DBN at time $t$ given the observation sequence up to $t$ (Boyen & Koller, 1998). Thus we can ask the following question: how does the policy hierarchy affect the size of the belief state representation of the corresponding AHMM?

Generally, for a policy hierarchy with $K$ levels, the belief state would have at least $K$ variables and thus the size of their joint distribution would be $O(exp(K))$. However, the AHMM has a specific network structure that exhibits certain conditional independence properties among its variables which can be exploited for efficiency. We first identify these useful independence properties in the AHMM and show that there is a compact representation of the special belief state in the case where the state sequence can be correctly observed (full observability assumption) and the starting and ending time of each policy is known. Consequently, policy recognition in this case can be performed very efficiently by updating the AHMM compact belief state. This partial result, although too restricted to be useful by itself, leads to an important observation about the general belief state: although it cannot be represented compactly, it can be approximated efficiently by a collection of compact special belief states. This makes the inference problem in the AHMM particularly amenable to a technique called *Rao-Blackwellisation* (Casella & Robert, 1996) which allows us to construct hybrid inference methods that combine both exact inference and approximate sampling-based inference for greater efficiency. The application of Rao-Blackwellisation to the AHMM structure reduces the sampling space that we need to approximate to a space with fixed dimension that does not depend on $K$, ensuring that the hybrid inference algorithm scales well w.r.t $K$.

The contributions of the paper are thus twofold. In terms of stochastic processes and dynamic Bayesian networks, we introduce the AHMM, a novel type of stochastic processes, provide its DBN structure and analyse the properties of this network. We present an application of the Rao-Blackwellised Particle Filter to the AHMM which results in an efficient hybrid inference method for this stochastic model. In terms of plan recognition, we propose a novel plan recognition framework based on probabilistic inference using the AHMM as the plan execution model. The complexity of the inference problem is addressed by applying a range of recently developed techniques in probabilistic reasoning to the plan recognition problem. Our work illustrates that while the stochastic models for plan execution can be complex, they exhibit certain special structures that can be exploited to construct efficient plan recognition algorithms.

### 1.2 Structure of the Paper

The main body of the paper is organised as follows. Section 2 introduces the background material in dynamic Bayesian networks and probabilistic inference. Section 3 formally defines the abstract Markov policy and the policy hierarchy. Section 4 presents the AHMM, its DBN representation and conditional independence properties. The algorithms for policy recognition are discussed in Section 5, first for the special tractable case and then for the general case. Section 6 presents our experimental results with the AHMM framework, including a real-time system for recognising people behaviour in a complex spatial environment using distributed video surveillance data. Section 7 provides a comparative review of related work in probabilistic plan recognition. Finally, we conclude and discuss directions for further research in Section 8.

## 2. Background in Probabilistic Inference

The aim of this section is to familiarise readers with some concepts in probabilistic inference that will be used later on in the paper. In subsections 2.1 and 2.2, we discuss Bayesian Networks (BN) and Dynamic Bayesian Networks (DBN) in general. In subsection 2.3, we discuss the Sequential Importance Sampling (SIS) algorithm, a general approximate sampling-based inference method for dynamic models. Subsections 2.4 and 2.5 introduce Rao-Blackwellisation, a technique for improving sampling-based methods by utilising certain special structures of the dynamic model. Later on, Rao-Blackwellisation will be used as our key computational technique for performing policy recognition.

### 2.1 Bayesian Networks

The *Bayesian network* (BN) (Pearl, 1988; Jensen, 1996; Castillo, Gutierrez, & Hadi, 1997) (also known as *probabilistic network* or *belief network*) is a well-established framework for dealing with uncertainty. It provides a graphical and compact representation of the joint probability distribution of a set of domain variables $X_1, \dots X_n$ in the form of a directed acyclic graph (DAG) whose nodes correspond to the domain variables. For each node $X_i$, the links from the parent nodes $Pa(X_i)$ are parameterised by the conditional probability of that node given the parents $\Pr(X_i \mid Pa(X_i))$. The network structure together with the parameters encode a factorisation of the joint probability distribution (JPD) $\Pr(X_1, \dots X_n) = \prod_{i=1}^{n} \Pr(X_i \mid Pa_i)$. Given a Bayesian network, conditional independence statements of the form $X \perp Y \mid Z$ (X is independent of Y given Z, where $X, Y, Z$ are variables or sets of variables) can be asserted if $X$ is d-separated from $Y$ by $Z$ in the network structure, where d-separation is a graph separation concept for DAGs (Pearl, 1988). The network structure of a BN thus captures certain conditional independence properties among the domain variables which can be exploited for efficient inference.

The main inference task on a Bayesian network is to calculate the conditional probability of a set of variables given the values of another set of variables (the evidence). There are two types of computation techniques for doing this. Exact inference algorithms (Lauritzen & Spiegelhalter, 1988; Jensen, Lauritzen, & Olesen, 1990; D'Ambrosio, 1993) compute the exact value of the conditional probability required based on analytical transformation that exploits the conditional independence relationships of the variables in the network.

Approximative inference algorithms (Pearl, 1987; York, 1992; Henrion, 1988; Fung & Chang, 1989; Shachter & Peot, 1989) compute only an approximation of the required probability, usually obtained either through "forward" sampling (Henrion, 1988; Fung & Chang, 1989; Shachter & Peot, 1989) (a variance of Bayesian Importance Sampling (Geweke, 1989)), or through Gibbs (Monte-Carlo Markov-Chain) sampling (Pearl, 1987; York, 1992). These algorithms have the advantages of simple implementation, can be applied to all types of network, and can trade off the accuracy in the estimates for computation resources. It is known that exact inference in BN is NP-hard with respect to the network size (Cooper, 1990), while approximate inference, although scales well with the network size, is NP-hard with respect to the hard-bound accuracy of the estimates (Dagum & Luby, 1993). In the light of these theoretical results, approximate inference can be useful in large networks when exact computation is intractable, but a certain degree of error in the probability estimate can be tolerated by the application.

### 2.2 Dynamic Bayesian Networks

To model the temporal dynamics of the environment, the Dynamic Bayesian Network (DBN) (Dean & Kanazawa, 1989; Nicholson & Brady, 1992; Dagum, Galper, & Horvitz, 1992) is a special Bayesian network architecture for representing the evolution of the domain variables over time. A DBN consists of a sequence of time-slices where each time-slice contains a set of variables representing the state of the environment at the current time. A time-slice is in itself a Bayesian network, with the same network structure replicated at each time-slice. The temporal dynamics of the environment is encoded via the network links from one time-slice to the next. In addition, each time-slice can contain observation nodes which model the (possibly noisy) observation about the current state of the environment.

Given a DBN and a sequence of observations, we might want to draw predictions about the future state variables (predicting), or about the unobserved variables in the past (smoothing) (Kjaerulff, 1992). This problem can be solved using an inference algorithm for Bayesian networks described above. However, if we want to revise the prediction as the observations arrive over time, reapplying the inference algorithm each time the observation sequence changes could be costly, especially as the sequence grows. To avoid this, we need to keep the joint distribution of all the variables in the current time-slice, given the observation sequence up to date. This probability distribution is termed the *belief state* (also known as the *filtering distribution*) and plays an important role in inferencing in the DBN. All existing inference schemes for the DBN involve maintaining and updating the belief state (i.e., filtering). When a new observation is received, the current belief state is rolled over one time-slice ahead following the evolution model, then conditioned on the new observation to obtain the updated belief state.

An obvious problem with this approach is the size of the belief state that we need to maintain. It has been noted that while the interaction of the variables in the DBN is localised, the variables in the belief state can be highly connected (Boyen & Koller, 1998). This is because the marginalisation of the past time-slices usually destroys the conditional independence of the current time-slice. When the size of the belief state is large, exact inference methods like (Kjærulff, 1995) is intractable, and it becomes necessary to maintain only an approximation of the actual belief state, either in the form of an approximate distribution that can be represented compactly (Boyen & Koller, 1998), or in the form of a set of weighted samples as in the Sequential Monte-Carlo Sampling methods (Doucet, Godsill, & Andrieu, 2000b; Kanazawa, Koller, & Russell, 1995; Liu & Chen, 1998).

The most simple case of the DBN where, in each time-slice, there is only a single state variable and an observation node, is the well-known Hidden Markov Model (HMM) (Rabiner, 1989). Filtering in this simple structure can be solved using dynamic programming in the discrete HMM (Rabiner, 1989), or Kalman filtering in the linear Gaussian model (Kalman, 1960). More recently, extensions of the HMM with multiple hidden interacting chains such as the Coupled Hidden Markov Models (CHMM) and the Factorial Hidden Markov Models (FHMM) have been proposed (Brand, 1997; Ghahramani & Jordan, 1997; Jordan, Ghahramani, & Saul, 1997). In these models, the size of the belief state is exponential in the number of hidden chains. Therefore, the inference and parameter estimation problems become intractable if the number of hidden chains is large. For this reason, approximate techniques are required. CHMM (Brand, 1997) employs a deterministic approximation that approximates full dynamic programming by keeping only a fixed number of "heads" with highest probabilities. The "heads" are thus chosen deterministically rather than randomly as in sampling-based methods. FHMM (Ghahramani & Jordan, 1997; Jordan et al., 1997) uses variational approximation (Jordan, Ghahramani, Jaakkola, & Saul, 1999) which approximates the full FHMM structure by a sparsified tractable structure. This idea is similar to the structured approximation method in (Boyen & Koller, 1998).

Our AHMM can be viewed as a type of Coupled/Factorial HMM since the AHMM also consists of a number of interacting chains. However the type of interaction in our AHMM is different from the other types of interaction that have been considered (Brand, 1997; Jordan et al., 1997; Ghahramani & Jordan, 1997). This is because the main focus of the AHMM is the dynamics of temporal abstraction among the chains, rather than the correlation between them at the same time interval. In addition, each node in the AHMM has a specific meaning (policy, state, or policy termination status), and the links have a clear causal interpretation based on the policy selection and persistence model. This is in contrast to the Coupled/Factorial HMM where the nodes and links usually do not have any clear semantic/causal interpretation. The advantage is that prior knowledge about the temporal decomposition of an abstract process can be incorporated in the AHMM more naturally.

### 2.3 Sequential Importance Sampling (SIS)

Sequential Importance Sampling (SIS) (Doucet et al., 2000b; Liu & Chen, 1998), also known as Particle Filter (PF), is a general Monte-Carlo approximation scheme for dynamic stochastic models. In principle, the SIS method is the same as the so-called Bayesian Importance Sampling (BIS) estimator in the static case (Geweke, 1989). Suppose that we want to estimate the quantity $\bar{f} = \int f(x)p(x)dx$, i.e., the mean of $f(x)$ where $x$ is a random variable with density $p$. Note that if $f$ is taken as the identity function of an event $A$ then $\bar{f}$ is simply $\Pr(A)$. Let $q(x)$ be an arbitrary[2] density function, termed the *importance distribution*. Usually, the importance distribution $q$ is chosen so that is it easy to obtain

2. For the weight to be properly defined, the support of $q$ has to be a subset of the support of $p$.

random samples from it. The expectation under estimation can then be rewritten as:

$$\bar{f} = \frac{\int [f(x)p(x)/q(x)]q(x)dx}{\int [p(x)/q(x)]q(x)dx} = \frac{\mathrm{E}_q\, f(x)p(x)/q(x)}{\mathrm{E}_q\, p(x)/q(x)}$$

From this expression, the BIS estimator w.r.t $q$ can be obtained:

$$\bar{f} \approx \hat{f}_{BIS} = \frac{\frac{1}{N}\sum_{i=1}^{N} f(x^{(i)})\dot{w}(x^{(i)})}{\frac{1}{N}\sum_{i=1}^{N} \dot{w}(x^{(i)})} = \sum_{i=1}^{N} f(x^{(i)})\tilde{w}(x^{(i)})$$

where $\{x^{(i)}\}$ are the $N$ i.i.d samples taken from $q(x)$, $\dot{w}(x) = p(x)/q(x)$ and $\tilde{w}$ is the normalised weight $\tilde{w}(x^{(i)}) = \dot{w}(x^{(i)})/\sum_i \dot{w}(x^{(i)})$. Note that the normalised weight can be computed from any weight function $w(x) \propto \dot{w}(x)$, i.e., the weight function need only be computed up to a normalising constant factor.

In the dynamic case, we want to estimate $\bar{f} = \int_{\tilde{x}_t} f(\tilde{x}_t)p(\tilde{x}_t|\tilde{o}_t)$ where $\tilde{x}_t = (x_0, \ldots, x_t)$ and $\tilde{o}_t = (o_0, \ldots, o_t)$ are two sequences of random variables; $o_t$ represents the observation available to us at time $t$. Often, $(\tilde{x}_t)$ is a Markov sequence and $o_t$ is the observation of $x_t$ as in a HMM. In a DBN, $x_t$ corresponds to the set of state variables and $o_t$ corresponds to the set of observations at time-slice $t$. The SIS method presented here however applies to the most general case where $(\tilde{x}_t)$ can be non-Markov, and $o_t$ not only depends on $x_t$.

We now can introduce the importance distribution $q(\tilde{x}_t|\tilde{o}_t)$ to obtain the estimator:

$$\bar{f} \approx \hat{f}_{SIS} = \sum_{i=1}^{N} f(\tilde{x}_t^{(i)})\tilde{w}(\tilde{x}_t^{(i)}) \tag{1}$$

To ensure that we can obtain sample from $q(\tilde{x}_t|\tilde{o}_t)$ "online", i.e., to sample a new value $x_t$ for the sequence $\tilde{x}_t$ when the current observation $o_t$ arrives, $q$ must be restricted to the form:

$$q(\tilde{x}_t|\tilde{o}_t) = q(\tilde{x}_{t-1}|\tilde{o}_{t-1})q(x_t|\tilde{x}_{t-1}, \tilde{o}_t)$$

With this restriction on $q$, we can use the weight function $w(\tilde{x}_t) = p(\tilde{x}_t, \tilde{o}_t)/q(\tilde{x}_t|\tilde{o}_t)$ so that the weight can also be updated "online" using:

$$w(\tilde{x}_t) = w(\tilde{x}_{t-1})p(x_t, o_t|\tilde{x}_{t-1}, \tilde{o}_{t-1})/q(x_t|\tilde{x}_{t-1}, \tilde{o}_t) \tag{2}$$

Let $w_t = w(\tilde{x}_t)/w(\tilde{x}_{t-1})$ be the weight updating factor at time $t$, and $q_t = q(x_t|\tilde{x}_{t-1}, \tilde{o}_t)$ be the sampling distribution used at time $t$. From (2) we have

$$w_t q_t = p(x_t, o_t|\tilde{x}_{t-1}, \tilde{o}_{t-1}) \tag{3}$$

which means that $p(x_t, o_t|\tilde{x}_{t-1}, \tilde{o}_{t-1})$ is factorised into two parts: $w_t$ and $q_t$. By choosing different factorisations, we obtain different forms for $q_t$ and thus different important distributions $q$. For example, when $(\tilde{x}_t, \tilde{o}_t)$ is a HMM, $q_t$ can be chosen as $p(x_t|x_{t-1})$ with $w_t = p(o_t|x_t)$ as in the likelihood weighting (LW) method, or $q_t$ can be chosen as $p(x_t|x_{t-1}, o_t)$ with $w_t = p(o_t|x_{t-1})$ as in the likelihood weighting with evidence reversal (LW-ER) (Kanazawa et al., 1995). In general, the "forward" $q_t$ can be chosen as $p(x_t|\tilde{x}_{t-1}, \tilde{o}_{t-1})$ with the corresponding weight $w_t = p(o_t|\tilde{x}_t, \tilde{o}_{t-1})$. The "optimal" $q_t$, in the sense discussed in (Doucet et al., 2000b), is chosen as $q_t = p(x_t|\tilde{x}_{t-1}, \tilde{o}_t)$ with the associating $w_t = p(o_t|\tilde{x}_{t-1}, \tilde{o}_{t-1})$.

The general SIS approximation scheme is thus as follows. At time $t-1$, we maintain $N$ sample sequences $\{\tilde{x}_{t-1}^{(i)}\}$ and the $N$ corresponding weight values $\{w^{(i)}\}$. When the current observation $o_t$ arrives, each sequence $\tilde{x}_{t-1}^{(i)}$ is lengthened by a new value $x_t^{(i)}$ sampled from the distribution $q(x_t|\tilde{x}_{t-1}^{(i)}, \tilde{o}_t)$. The weight value for $\tilde{x}_t^{(i)}$ is then updated using (2). Once the new samples and the new weights are obtained, the expectation of any functional $f$ can be estimated using (1). This procedure can be furthered enhanced with a re-sampling step and a Markov-chain sampling step (see Doucet et al. (2000b), Doucet, de Freitas, Murphy, and Russell (2000a)). We do not describe these important improvements of the SIS here.[3]

### 2.4 Rao-Blackwellisation

Rao-Blackwellisation is a general technique for improving the accuracy of sampling methods by analytically marginalising some variables and only sampling the remainder (Casella & Robert, 1996). In its simplest form, consider the problem of estimating the expectation $\mathrm{E}\, f(x)$, where $x$ is a joint product of two variables $r, z$. Using direct Monte-Carlo sampling, we obtain the estimator: $\hat{f} = \frac{1}{N}\sum_1^N f(r^{(i)}, z^{(i)})$. Alternatively, a Rao-Blackwellised estimator can be derived by sampling only the variable $r$, with the other variable $z$ being integrated out analytically:

$$\mathrm{E}\, f(r,z) = \mathop{\mathrm{E}}_r h(r) \approx \hat{f}_{RB} = \frac{1}{N}\sum_1^N h(r^{(i)})$$

where $h(r) = \mathrm{E}_z[f(r,z)|r]$. For our convenience, $r$ will be referred to as the *Rao-Blackwellising variable.*

The Rao-Blackwellised estimator $\hat{f}_{RB}$ is generally more accurate than $\hat{f}$ for the same number of samples $N$. This is a direct consequence of the Rao-Blackwell theorem which gives the relationship between unconditional and conditional variance:

$$\mathrm{VAR}\, X = \mathrm{VAR}[\mathrm{E}[X|Y]] + \mathrm{E}[\mathrm{VAR}[X|Y]]$$

When applying to the problem of estimating $\mathrm{E}\, f(r,z)$, we have:

$$\mathrm{VAR}\, f(r,z) = \mathrm{VAR}[\mathrm{E}[f(r,z)|r]] + \mathrm{E}[\mathrm{VAR}[f(r,z)|r]]$$

and thus $\mathrm{VAR}\, f(r,z) \geq \mathrm{VAR}[\mathrm{E}[f(r,z)|r]] = \mathrm{VAR}\, h(r)$. This suggests that for direct Monte-Carlo sampling, the error of RB-sampling (sample only $r$ and marginalise $z$) is always smaller than the error of sampling both $r$ and $z$ for the same number of samples, except in the degenerated case. For Bayesian Importance Sampling, using the variance convergence result from (Geweke, 1989), one can also easily prove that as the number of samples tend to infinity, the RB-BIS would generally do better than BIS for the same number of samples.

3. Note that these improvements can be used orthogonal to the Rao-Blackwellisation procedure discussed subsequently. Our implementation of the policy recognition algorithm in the later sections does include a re-sampling step, which is crucial for keeping the error of SIS over time under control.

### 2.5 SIS with Rao-Blackwellisation (RB-SIS)

Since SIS is a form of BIS, Rao-Blackwellisation can also be used to improve its performance (Liu & Chen, 1998; Doucet et al., 2000b). Let us consider again the problem of estimating the expectation $\bar{f} = \int f(\tilde{x}_t)p(\tilde{x}_t|\tilde{o}_t)$, where each variable $x_t$ is the joint product of two variables $(z_t, r_t)$. We shall restrict ourselves to the case where $\tilde{x}_t$ is Markov and $o_t$ is an observation of $x_t$, i.e., when $(\tilde{x}_t, \tilde{o}_t)$ can be represented by a DBN. In addition, we only consider $f$ that depends only on the current variable $x_t$, i.e., $\bar{f}$ is an expectation over the filtering distribution $p(x_t|\tilde{o}_t)$. For example, if $A$ is a "future" event, i.e., an event that depends on $\{x_{t'}|t' \geq t\}$, we can estimate $p(A|\tilde{o}_t)$ by letting $f(x_t) = p(A|x_t)$ so that $\bar{f} = \int_{x_t} p(A|x_t)p(x_t|\tilde{o}_t) = p(A|\tilde{o}_t)$.

Applying Rao-Blackwellisation to this setting, we can let $h(\tilde{r}_t) = \int_{z_t} f(z_t, r_t)p(z_t|\tilde{r}_t, \tilde{o}_t)$, so that $\bar{f} = \bar{h} = \int_{\tilde{r}_t} h(\tilde{r}_t)p(\tilde{r}_t|\tilde{o}_t)$. Thus, if we use SIS to estimate $\bar{h}$, we also obtain an estimator for $\bar{f}$:

$$\bar{f} \approx \hat{f}_{RBSIS} = \hat{h}_{SIS} = \sum_{i=1}^{N} h(\tilde{r}_t^{(i)})\tilde{w}(\tilde{r}_t^{(i)}) \tag{4}$$

The benefit of doing this is the increase in the accuracy of the estimator, as we now only need to sample the variables $\tilde{r}_t$. The down side is that for each sample $\tilde{r}_t$, we need to compute $h(\tilde{r}_t)$ using some exact inference method. Furthermore, the SIS procedure to estimate $\bar{h}$ might require some additional complexity since the sequence $\tilde{r}_t$ is generally non-Markov, and $o_t$ no longer depends only on $r_t$. Overall, in comparison with the normal SIS estimator $\hat{f}_{SIS}$ (Eq. 1), for the same number of samples $N$, $\hat{f}_{RBSIS}$ is more accurate but is also more computationally demanding to compute.

To see more clearly what is involved in implementing the RB-SIS method, let us look at the *Rao-Blackwellised belief state*, i.e., the belief state of the dynamic process when the Rao-Blackwellising variables can be observed: $\mathcal{R}_t = p(z_t, r_t, o_t|\tilde{r}_{t-1}, \tilde{o}_{t-1})$ and its posterior $\mathcal{R}_{t+} = p(z_t|\tilde{r}_t, \tilde{o}_t)$. All the entities needed in the RB-SIS procedure can be computed from these two distributions. Indeed, the functional $h$ can be rewritten in terms of $\mathcal{R}_{t+}$ as:

$$h(\tilde{r}_t) = \int_{z_t} f(z_t, r_t)p(z_t|\tilde{r}_t, \tilde{o}_t) = \int_{z_t} f(z_t, r_t)\mathcal{R}_{t+}(z_t) \tag{5}$$

In addition, while performing SIS to estimate $\bar{h}$, from Eq. (3), the weight $w_t$ and the sampling distribution $q_t$ can be computed from $\mathcal{R}_t$:

$$w_t q_t = p(r_t, o_t|\tilde{r}_{t-1}, \tilde{o}_{t-1}) = \mathcal{R}_t(r_t, o_t) = \int_{z_t} \mathcal{R}_t(z_t, r_t, o_t) \tag{6}$$

Thus, computing the RB belief state $\mathcal{R}_t$ and its posterior $\mathcal{R}_{t+}$ is an essential step in the RB-SIS method. Since we have to maintain an RB belief state for each sample of the RB variables $\tilde{r}_t$, it is crucial that this can be done efficiently using an exact inference method. If $x_t$ is composed of many variables, as in the case of a DBN, our choice of the Rao-Blackwellising variables should be so that the Rao-Blackwellised belief state can be maintained in a tractable way. Hence, Rao-Blackwellisation is especially useful when the set of variables in a DBN can be split into two parts such that conditioning on the first part makes the structure of the second part tractable and amenable to exact inference.

Begin
For $t = 0, 1, \ldots$
  For each sample $i = 1, \ldots, N$
    Sample $r_t^{(i)}$ from $\mathcal{R}_t^{(i)}(r_t|o_t)$
    Update weight $w^{(i)} = w^{(i)}\mathcal{R}_t^{(i)}(o_t)$
    Compute the posterior RB bel state $\mathcal{R}_{t+}^{(i)} = \mathcal{R}_t^{(i)}(z_t|r_t^{(i)}, o_t)$
    Compute the new RB belief state $\mathcal{R}_{t+1}^{(i)}$ from $\mathcal{R}_{t+}^{(i)}$
    Compute $h^{(i)}$ from $\mathcal{R}_{t+}^{(i)}$
  Compute the estimator $\hat{f}_{RBSIS} = \sum_{i=1}^{N} h^{(i)}\tilde{w}^{(i)}$
End

Figure 1: RB-SIS for general DBN

The general RB-SIS algorithm is given in Fig. 1. For illustrating purpose, we assume that the "optimal" $q_t$ and the corresponding $w_t$ are being used ($q_t = \mathcal{R}_t(r_t|o_t)$ and $w_t = \mathcal{R}_t(o_t)$). At each time point, we need to maintain $N$ samples $\tilde{r}_t^{(i)}$, $i = 1, \ldots, N$. For each sample, in addition to the sample weight $w^{(i)}$, we also need to store a representation of the RB belief state corresponding to that sample sequence: $\mathcal{R}_t^{(i)} = p(r_t, z_t, o_t|\tilde{r}_{t-1}^{(i)}, \tilde{o}_{t-1})$ and $\mathcal{R}_{t+}^{(i)} = p(z_t|\tilde{r}_t^{(i)}, \tilde{o}_t)$.

A number of applications of the RB-SIS method (also known as the Rao-Blackwellised Particle Filter (RBPF)) have been discussed in the literature. A general framework for using RB-SIS to do inference on DBNs has been presented by Doucet et al. (2000a), Murphy (2000), Murphy and Russell (2001). However, these authors have mainly focused on the case where the sequence of the Rao-Blackwellising variables ($\tilde{r}_t$) is Markov (for example, when the RB variables are the root nodes at each time slice). This assumption simplifies the sampling step in the RB procedure since obtaining the sample for the RB variable at time $t+1$ is straightforward. In our previous work (Bui, Venkatesh, & West, 2000), we introduced a hybrid-inference method for the AHMM in the special case of the state-space decomposition policy hierarchy, which is essentially an RB-SIS method. Note that when applied to AHMMs, the sequence of Rao-Blackwellising variables that we use does not satisfy the Markov property. In this case, care must be taken to design an efficient sampling step, especially when the sampling distribution for the next RB variable does not have a tractable form. The use of non-Markov RB variables also appears in other special models such as the Bayesian missing data model (Liu & Chen, 1998), and the partially observed Gaussian state space model (Andrieu & Doucet, 2000) where the RB belief state can be maintained by a Kalman filter.

Since we have to make the Rao-Blackwellised belief state tractable, the context variables in the framework of context-specific independence (Boutilier, Friedman, Goldszmidt, & Koller, 1996) can be used conveniently as Rao-Blackwellising variables (Murphy, 2000). Indeed, since the context variable acts as a mixing gate for different Bayesian network structures, conditioning on these variables would simplify the structure of the remaining vari-ables. Because of this property of the context variables, Boutilier et al. (1996) have suggested to use them as the cut-set variables in the cut-set conditioning inference method (Pearl, 1988). The cut-set variables play a similar role to the Rao-Blackwellising variables in which they help to simplify the structure of the remaining network. In Rao-Blackwellised sampling, instead of summing over all the possible values of the cut-set variables which can be intractable, only a number of representative sampled values are used.

The idea of combining both exact and approximate inference in RB sampling is also similar to the hybrid inference scheme described by Dawid, Kjærulff, and Lauritzen (1995), however it's unclear if RB sampling can be described using their model of communicating belief universe. Also, Dawid et al. use hybrid inference mainly to do inference on networks with a mixture of continuous and discrete variables, as opposed to RB whose goal is to improve the sampling performance.

## 3. Abstract Markov Policies

In this section, we formally introduce the AMP concept as originating from the literature of abstract probabilistic planning with MDPs (Sutton et al., 1999; Parr & Russell, 1997; Forestier & Varaiya, 1978; Hauskrecht et al., 1998; Dean & Lin, 1995). The main motivation in abstract probabilistic planning is to scale up MDP-based planning to problems with large state space. It has been noted that a hierarchical organisation of policies can help reduce the complexity of MDP-based planning, similar to the role played by the plan hierarchy in classical planning (Sacerdoti, 1974). In comparison with a classical plan hierarchy, a policy hierarchy can model different sources of uncertainty in the planning process such as stochastic actions, uncertain action outcomes, and stochastic environment dynamics.

While the work in planning is concerned with finding the optimal policy given some reward function, our work focuses on policy recognition which is the inverse problem, i.e., to infer the agent's policies from watching the effects of the agent's actions. The two problems however share a common element which is the model of a stochastic plan hierarchy. In policy recognition, although it is possible to derive some information about the reward function by observing the agent's behaviour, we choose not to do this, thus omitting from our model the reward function and also the optimality notion. This leaves the model open to tracking arbitrary agent's behaviours, regardless of whether they are optimal or not.

### 3.1 The General Model

#### 3.1.1 Actions and Policies

In an MDP, the world is modelled as a set of possible states $S$, termed the state space. At each state $s$, an agent has a set of actions $A$ available, where each action $a$, if employed, will cause the world to evolve to the next state $s'$ via a transition probability $\sigma_a(s, s')$. An agent's plan of actions is modelled as a policy that prescribes how the agent would choose its action at each state. For a policy $\pi$, this is modelled by a selection function $\sigma_\pi : S \times A \to [0, 1]$ where at each state $s$, $\sigma_\pi(s, a)$ is the probability that the agent will choose the action $a$. It is easy to see that, given a fixed policy $\pi$, the resulting state sequence is a Markov chain with transition probabilities $\Pr(s' \mid s) = \sum_a \sigma_\pi(s, a)\sigma_a(s, s')$. Thus, a policy can also be viewed as a Markov chain through the state space.

### 3.1.2 Local Policies

In the original MDP, behaviours are modelled at only two levels: the primitive action level, and the plan level (policy). We would like to consider policies that select other more refined policies and so on, down a number of abstraction levels. The idea is to form intermediate-level abstract policies as policies defined over a local region of the state space, having a certain terminating condition, and can be invoked and executed just like primitive actions (Forestier & Varaiya, 1978; Sutton et al., 1999).

**Definition 1 (Local policy).** A local policy is a tuple $\pi = \langle S, D, \beta, \sigma \rangle$ where:

- $S$ is the set of applicable states.
- $D$ is the set of destination states. $\beta : D \to (0, 1]$ is the stopping probabilities such that $\beta(d) = 1, \forall\, d \in D \setminus S$.
- $\sigma : S \times A \to [0, 1]$ is the selection function. Given the current state $s$, $\sigma(s, a)$ is the probability that the action $a$ is selected by the policy $\pi$ at state $s$.

The set $S$ models the local region over which the policy is applicable. $S$ will be called the set of applicable states, since the policy can start from any state in $S$. We shall assume here that $S$ is discrete, and thus shall not be concerned with the technical details in generalising the AHMM formulation to the continuous state space case. The stopping condition of the policy is modelled by a set of possible destination states $D$ and a set of positive stopping probabilities $\beta(d), d \in D$ where $\beta(d)$ is the probability that the policy will terminate when the current state is $d$. It is possible to allow the policy to stop at some state outside of $S$, however, for all $d \in D \setminus S$ we enforce the condition that $\beta(d) = 1$, i.e., $d$ is a terminal destination state. Sometimes, we might only want to consider policies with deterministic stopping condition. In that case, every destination is a terminal destination: $\forall d \in D$, $\beta(d) = 1$. Thus, for a deterministically terminating policy, we can ignore the redundant parameter $\beta$, and need only specify the set of destinations $D$.

Given a starting state $s \in S$, a local policy as defined above generates a Markov sequence of states according to its transition model. Each time a destination state $d \in D$ is reached, the process stops with probability $\beta(d)$. Since the process starts from within $S$, but terminates only in one of the states in $D$, the destination states play the role of the possible exits out of the local region $S$ of the state space.

When we want to make clear which policy is currently being referred to, we shall use the subscripted notations $S_\pi$, $D_\pi$, $\beta_\pi$, $\sigma_\pi$ to denote the elements of the policy $\pi$.

Fig. 2 illustrates how a local policy $\pi$ can be visualised. Fig. 2(a) shows the set of applicable states $S$, the set of destinations $D$, and a chain starting within $S$ and terminating in $D$. The Bayesian network in Fig. 2(b) provides the detailed view of the chain from start to finish. The Bayesian network in Fig. 2(c) is the abstract view of the chain where we are only interested in its starting and stopping states.

### 3.1.3 Abstract Policies

The local policy as defined above selects among the set of primitive actions. Similarly, but more generally, we can define higher level policies that select among a set of other policies.

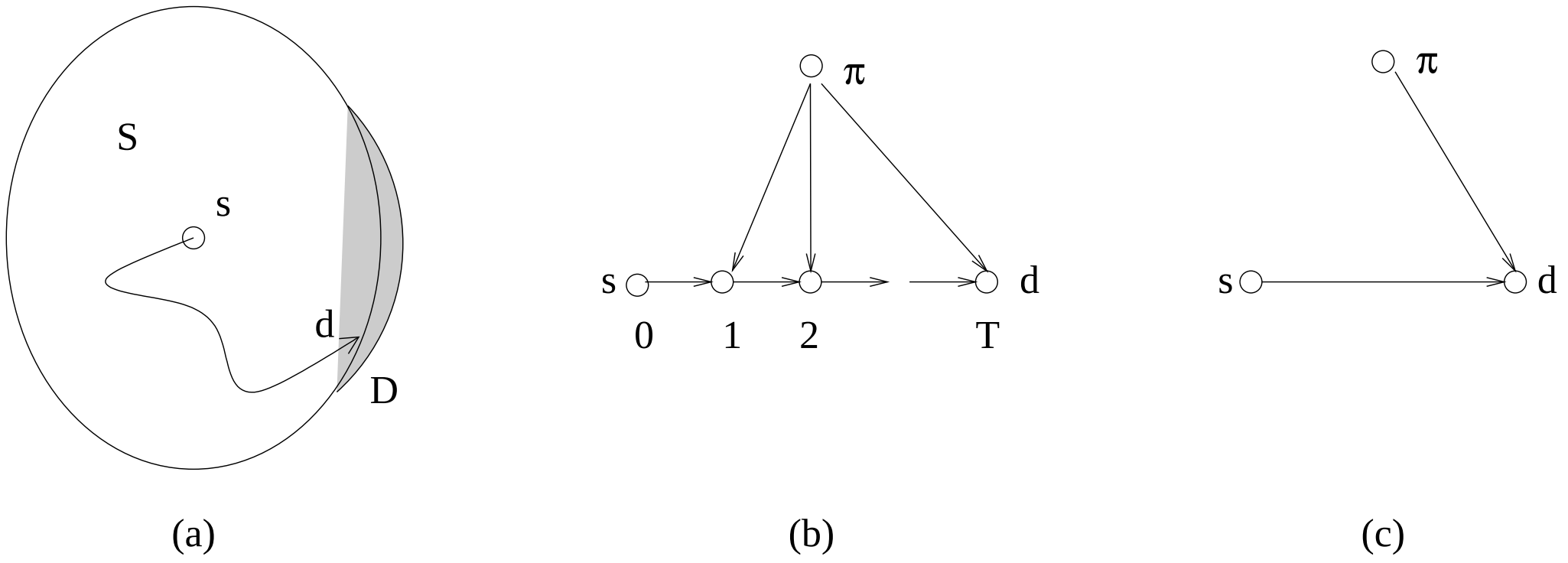

Figure 2: Visualisation of a policy

**Definition 2 (Abstract Policy).** Let $\Pi$ be a set of abstract policies. An abstract policy $\pi^*$ over the policies in $\Pi$ is a tuple $\langle S_{\pi^*}, D_{\pi^*}, \beta_{\pi^*}, \sigma_{\pi^*} \rangle$ where:

- $S_{\pi^*} \subset \cup_{\pi \in \Pi} S_\pi$ is the set of applicable states.
- $D_{\pi^*} \subset \cup_{\pi \in \Pi} D_\pi$ is the set of destination states. $\beta_{\pi^*} : D_{\pi^*} \to (0, 1]$ is the set of stopping probabilities.
- $\sigma_{\pi^*} : S_{\pi^*} \times \Pi \to [0, 1]$ is the selection function where $\sigma_{\pi^*}(s, \pi)$ is the probability that $\pi^*$ selects the policy $\pi$ at the state $s$.

Note the recursiveness in definition 2 that allows an abstract policy to select among a set of other abstract policies. At the base level, primitive actions are viewed as abstract policies themselves. Since primitive actions always stop after one time-step, $D_a \supset S_a$ and $\beta(d) = 1 \, \forall d \in D_a$ (Sutton et al., 1999). The idea that policies with suitable stopping condition can be viewed just as primitive actions is first made explicit in (Sutton, 1995), which also introduces the $\beta$ model for representing the stopping probabilities. Their subsequent work (Sutton et al., 1999) introduces the abstract policy concept under the name *options*.

The execution of an abstract policy $\pi^*$ is as follows. Starting from some state $s$, $\pi^*$ selects a policy $\pi \in \Pi$ according to the distribution $\sigma_{\pi^*}(s, .)$. The selected policy $\pi$ is then executed until it is terminated in some state $d \in D_\pi$. If $d$ is also a destination state of $\pi^*$ ($d \in D_{\pi^*}$), the policy $\pi^*$ stops with probability $\beta_{\pi^*}(d)$. If $\pi^*$ still continues, a new policy $\pi' \in \Pi$ is selected by $\pi^*$ at $d$, which will be executed until its termination and so on (Fig. 3).

Some remarks about the representation of an abstract policy are needed here. Let $s \in \cup_{\pi \in \Pi} S_\pi$, we denote the subset of policies in $\Pi$ which are applicable at $s$ by $\Pi(s) = \{\pi \in \Pi \,|\, s \in S_\pi\}$. For an abstract policy $\pi^*$ to be well-defined, we have to make sure that at each state $s$, $\pi^*$ only selects among the policies that are applicable at $s$. Thus, the selection function has to be such that $\sigma_{\pi^*}(s, \pi) > 0$ only if $\pi \in \Pi(s)$. This helps to keep the specification of the selection function to a manageable size, even when the set of all policies $\Pi$ to be chosen from can be large. In addition, the specification of the selection function and the stopping probabilities can make use of factored representations (Boutilier, Dearden, & Goldszmidt, 2000) in the case where the state space is the composite of a set of relatively independent variables. This ensures that we still have a compact specification of the probabilities conditioned on the state variable, even though the state space can be of high dimension.

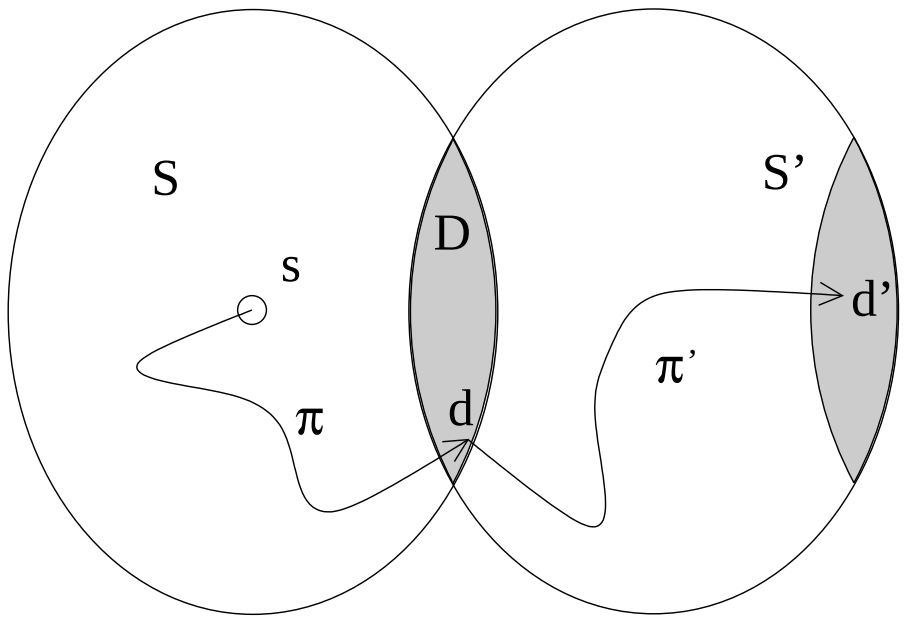

Figure 3: A chain generated by an abstract policy

#### 3.1.4 Policy Hierarchy

Using abstract policies as the building blocks, we can construct a hierarchy of abstract policies as follows:

**Definition 3 (Policy hierarchy).** A policy hierarchy is a sequence $\mathcal{H} = (\Pi_0, \Pi_1, \ldots, \Pi_K)$ where $K$ is the number of levels in the hierarchy, $\Pi_0$ is a set of primitive actions, and for $k = 1, \ldots, K$, $\Pi_k$ is a set of abstract policies over the policies in $\Pi_{k-1}$.

When a top-level policy $\pi^K$ is executed, it invokes a sequence of level-(K-1) policies, each of which invokes a sequence of level-(K-2) policies and so on. A level-1 policy will invoke a sequence of primitive actions which leads to a sequence of states. Thus, the execution of $\pi^K$ generates an overall state sequence $(s_0, s_1, \ldots, s_t, \ldots)$ that terminates in one of the destination states in $D_{\pi^K}$. When $K = 1$ this sequence is simply a Markov chain (with suitable stopping conditions). However, for $K \geq 2$, it will generally be non-Markovian, despite the fact that all the policies are Markov, i.e., they select the lower level policies based solely on the current state (Sutton et al., 1999). This is because knowing the current state $s_t$ alone does not provide information about the current intermediate-level policies, which can affect the selection of the next state $s_{t+1}$. Intuitively, this means that an agent's behaviour to achieve a given goal is usually non-Markovian, since its choice of actions depends not only on the current state, but also on the current intermediate intentions of the agent.

We term the dynamical process in executing a top-level abstract policy $\pi^K$ the *Abstract Markov Model* (AMM). When the states are only partially observable, the observation can be modelled by the usual observation model $\Pr(o_t \mid s_t) = \omega(s_t, o_t)$. The resulting process is termed the *Abstract Hidden Markov Model* (AHMM) since the states are hidden as in the Hidden Markov Model (Rabiner, 1989).

The idea of having a higher level policy controlling the lower level ones in an MDP can be traced back to the work by Forestier and Varaiya (1978), who investigated a two layer structure similar to our 2-level policy hierarchy with deterministic stopping condition. Forestier and Varaiya showed that that the sub-process, obtained by sub-sampling the state sequence at the time when the level-1 policy terminates, is also Markov, thus the policies at level 1 simply play the role of an "extended" action. In our framework, given a policy hierarchy, one can consider a "lifted" model where only the policies from level $k$ up and the observations at the time points when a policy at level $k$ ends are considered. The level-$k$ policies can then be considered as primitive actions, and the lifted model can be treated like a normal model.

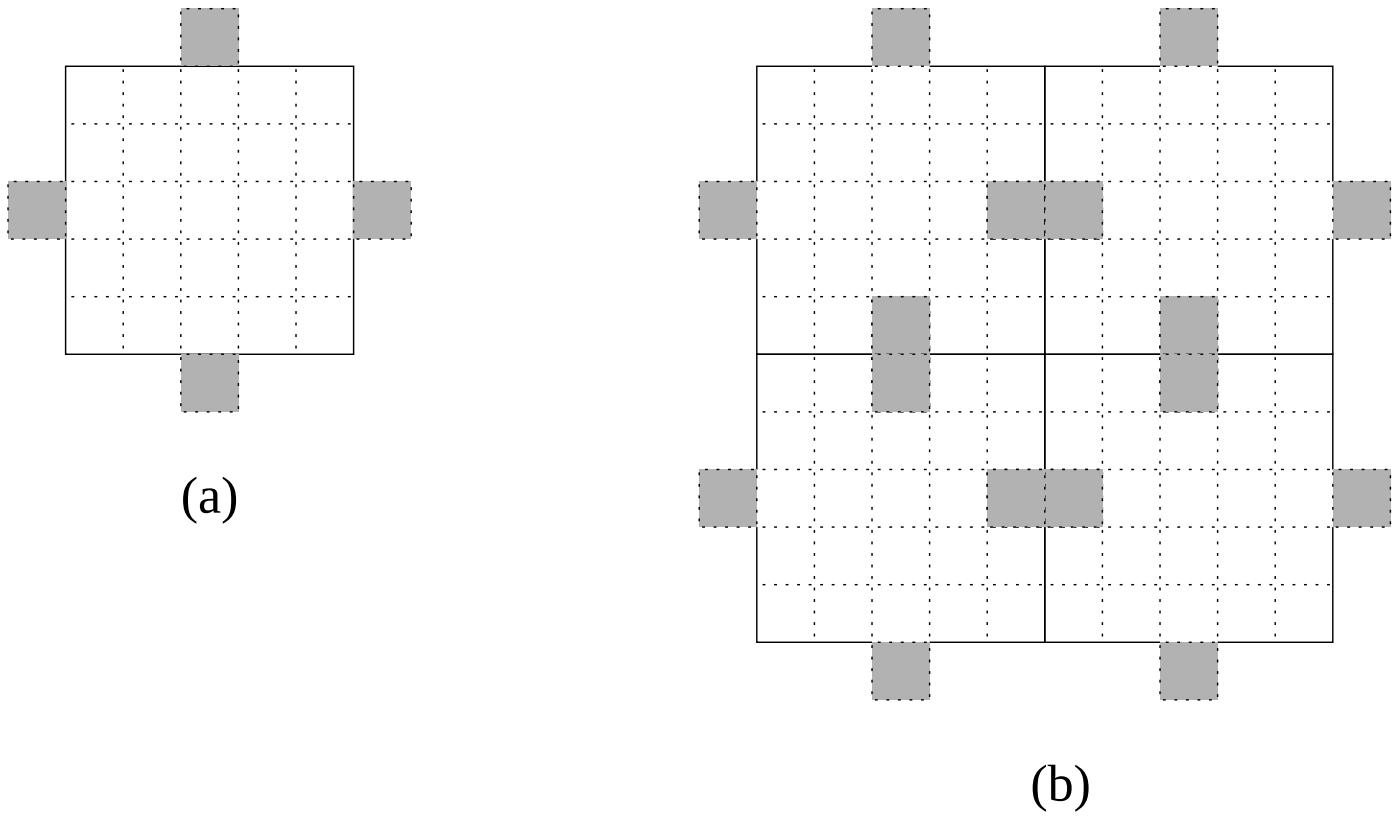

Figure 4: The environment and its partition

### 3.2 State-Space Region-Based Decomposition

In some cases, the state space or some of its dimensions already exhibit a natural hierarchical structure. For example, in the spatial domain, the set of ground positions can be divided into small local spaces such as rooms, corridors, etc. A set of these local spaces can be grouped together to form a larger space at the higher level (floors, buildings, etc). An intuitive and often-used method for constructing the policy hierarchy in this case is via the so-called region-based decomposition of the state space (Dean & Lin, 1995; Hauskrecht et al., 1998). Here, the state space $S$ is successively partitioned into a sequence of partitions $\mathcal{P}_K, \mathcal{P}_{K-1}, ...\mathcal{P}_1$ corresponding to the $K$ levels of abstraction, where $\mathcal{P}_K = \{S\}$ is the coarsest partition, and $\mathcal{P}_1$ is the finest. For each region $R_i$ of $\mathcal{P}_i$, the periphery of $R_i$, $Per(R_i)$ is defined as the set of states not in $R_i$, but connected to some state in $R_i$. Let $Per_i$ be the set of all peripheral states at level $i$: $Per_i = \cup_{R_i \in \mathcal{P}_i} Per(R_i)$. Fig. 4(b) shows an example where the state space representing a building is partitioned into 4 regions corresponding to the 4 rooms. The peripheral states for a region is shown in Fig 4(a), and Fig 4(b) shows all such peripheral states.

To construct the policy hierarchy, we first define for each region $R_1 \in \mathcal{P}_1$ a set of abstract policies applicable on $R_1$, and having $Per(R_1)$ as the destination states. For example, for each room in Fig 4, we can define a set of policies that model the agent's different behaviours while it is inside the room, e.g., getting out through a particular door. These policies can be initiated from inside the room, and terminate when the agent steps out of the room (not necessarily through the target door since the policy might fail to achieve its intended target). Note that since $Per(R_1) \cap R_1 = \emptyset$, all the policies defined in this manner have deterministic stopping conditions.

Let the set of all policies defined be $\Pi_1$. At the higher level $\mathcal{P}_2$, for each region $R_2$, we can define a set of policies that model the agent's behaviours inside that region with applicable state space $R_2$, destination set $Per(R_2)$, and the constraint that these policies must use the policies previously defined at level-1 to achieve their goals. An example is a policy to navigate between the room-doors to get from one building gate to another. Let the set of all policies defined at this level be $\Pi_2$. Continuing doing this at the higher levels, we obtain the policy hierarchy $\mathcal{H} = (\Pi_0, \Pi_1, \Pi_2, \ldots, \Pi_K)$. A policy hierarchy constructed through State-space Region-based Decomposition is termed an SRD policy hierarchy.

An SRD policy hierarchy has the property that the set of applicable states of all the policies at a given abstraction level forms a partition of the state space. Thus, from the state sequence $(s_0, \ldots, s_t, \ldots)$ resulting from the execution of the top level policy, we can infer the exact starting and terminating times of all intermediate-level policies. For example, at level $k$, the starting/stopping times of the policies in this level are the time indices $t$'s at which the state sequence crosses over a region boundary: $s_{t-1} \in R_k$ and $s_t \notin R_k$ for some region $R_k$ of the partition $\mathcal{P}_k$. Later in section 5.1, we will show that this property helps to simplify some of the complexity of the policy recognition problem.

### 3.3 A Policy Hierarchy Example

As an example, consider the task to monitor and predict the movement of an agent through a building shown in Fig. 5(a). Each room is represented by a $5 \times 5$ grid, and two adjacent rooms are connected via a door in the center of their common edge. The four entrances to the building are labeled north (N), west (W), south (S) and east (E). In addition, the door in the center of the building (C) acts like an entrance between the building's north wing and south wing. At each state (cell), the agent can move in 4 possible directions except when it is blocked by a wall.

The policy hierarchy to model the agent's behaviour in this environment can be constructed based on region-based decomposition at three levels of abstraction. Firstly, a region hierarchy is constructed. The partition of the environment consists of the 8 rooms at level 1, the two wings (north and south) at level 2, and the entire building at level 3. The behaviours of the agent at level 1 (within each room) is represented by a set of level 1 policies. For example, in each room, we use 4 level-1 policies to model the agent's behaviours of exiting the room via the 4 different doors. These are essentially four Markov chains within the room which terminate outside of the room. One way to represent these policies is to specify which movement action the agent should take given the current position and the current heading. At the higher level, the agent's behaviours within each wing are specified. For example, we use 3 level-2 policies in each wing to model the agent's behaviours of exiting the wing via the 3 wing exits. These policies are built on top of the set of level-1 policies already defined. They specify which level-1 policies the agent should take to leave the wing at the intended exit. Finally, at the top level, the agent's behaviours within the entire building can be specified. For example, we use 4 top-level policies to model the agent's behaviours of leaving the building via the four building exits N, W, S, E. A sample of these policies and their parameters is given in Fig. 5(b).

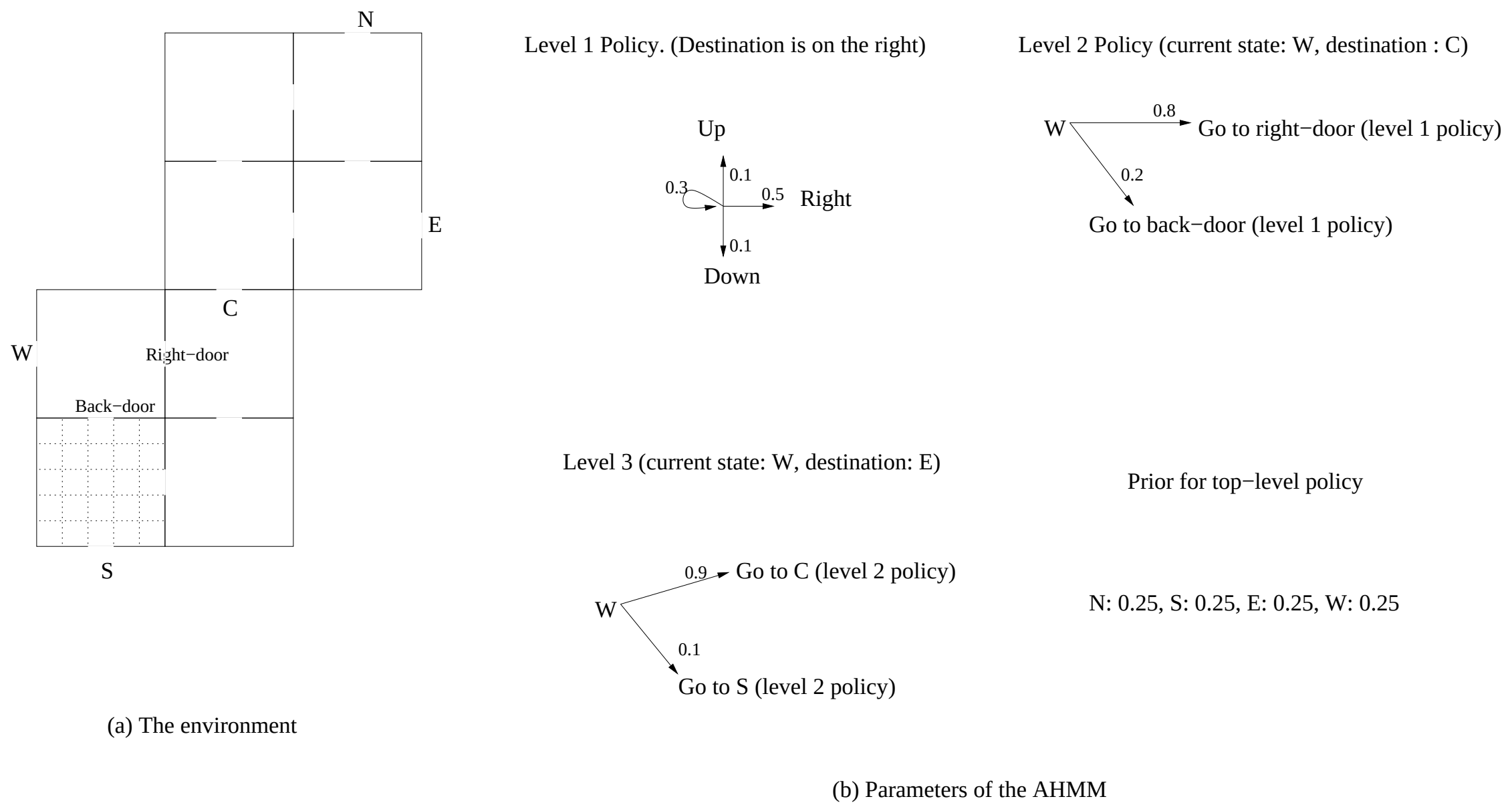

Figure 5: An example policy hierarchy

### 3.4 AMM as a Plan Execution Model

Up to now, we have presented the AMM as a formal plan execution model to be used later in the plan recognition process. In this subsection, we discuss the expressiveness of the AMM as a formal plan specification language, and also the suitability of using the AMM to encode plans in the context of plan recognition. Note that the discussion here focuses on the representational aspect of the AMM alone. A discussion of the computational aspects of the AMM/AHMM in comparison with other works in probabilistic plan recognition will be presented in Section 7.

The AMM is particularly well-suited for representing goal-directed behaviours at different levels of abstraction. Each policy in the AMM can be viewed as a plan trying to achieve a particular goal. However, unlike a classical plan, a policy specifies the course of actions at all applicable states, and is more similar to a *contingent plan*. The ending of a policy could either means that the goal has been achieved, or the attempt to achieve the goal using the current policy has failed. This interpretation of the persistence of a policy fits into the persistence model of *intentions* (Cohen & Levesque, 1990): when an intention ends, there is no guarantee that the intended goal has been achieved. Thus, conceptually, there are two types of destination states: one corresponds to the intended goal states, and the other corresponds to unintended failure states resulting from the stochastic nature in the execution of the plan. Due to its generality, the AMM does not need to distinguish between these two types; both the successful termination states and the unsuccessful ones are treated the same as possible destination states, albeit with different reaching probabilities.[4]

4. One would expect that an agent would more likely to reach the intended destination state rather a random failure state.

Using the AMM as a model of plan execution thus allows us to blur the difference between planning and re-planning. At the same time, it moves from the recognition of a classical plan towards the recognition of the agent's *intention*. Most of the existing framework for probabilistic plan recognition does not explicitly represent the current state, and thus, the relationship between states and the adoption and termination of current plans is ignored (Goldman et al., 1999).[5] Thus, it would be impossible to tell if the current plan has failed and the new plan is an attempt to recover from this failure, or the current plan has succeeded and the new plan is part of a new higher level goal.

A more expressive language for describing abstract probabilistic plan is the Hierarchical Abstract Machines (HAM) proposed in (Parr & Russell, 1997; Parr, 1998). In a HAM, the abstract policy is replaced by a stochastic finite automaton, which can call other machines at the lower level. Our abstract policies can be written down as machines of this type. Such a machine would choose one of the machines correspond to the policies at the lower level and then go back to the start state after the called machines have terminated. The HAM framework allows for machines with arbitrary finite number of machine states and transition probabilities,[6] thus can readily represent more complex plans such as concatenation of policies, alternative policy paths, etc. It is possible to represent each machine in HAM as a policy in our AMM, however with the cost of augmenting the state space to include the machine states of all the machines in the current call stack. Thus, the size of the AMM's new state space would be exponential with respect to the number of nested levels in the HAM's call stack. While this shows in theory the expressiveness of HAM and our policy hierarchy is the same, performing policy recognition on the HAM-equivalent policy hierarchy is probably unwise since the state space becomes exponentially large after the conversion. A better idea would be to represent the internal state of each machine as a variable in a DBN and perform inference on this DBN structure directly.

The AMM is also closely related to a model for probabilistic plan recognition called the Probabilistic State-Dependent Grammar (PSDG), independently proposed in (Pynadath, 1999; Pynadath & Wellman, 2000). The PSDG can be described as the Probabilistic Context Free Grammar (PCFG) (Jelinek, Lafferty, & Mercer, 1992), augmented with a state space, and a state transition probability table for each terminal symbol of the PCFG. In addition, the probability of each production rule is made state dependent. As a result, the terminal symbol now acts like primitive actions and the non-terminal symbol chooses its expansion depending on the current state. Interestingly, the PSDG is directly related to the HAM language described above, similar to the way production-rule grammars are related to finite automata. Given a PSDG, we can convert it to an equivalent HAM by constructing a machine for each non-terminating symbol, and modelling the production rules for each non-terminating symbol by the automaton.

Our policy hierarchy is equivalent to a special class of PSDG where only production rules of the form $X \to YX$ and $X \to \emptyset$ are allowed. The former rule models the adoption of a lower level policy $Y$ by a higher level policy $X$, while the latter models the termination of a policy $X$. The PSDG model considered in (Pynadath, 1999; Pynadath & Wellman, 2000) allows for more general rules of the form $X \to Y_1 \dots Y_m X$, i.e., the recursion symbol

5. with the exceptions of (Goldman et al., 1999; Pynadath & Wellman, 2000) which will be discussed in detail in Section 7.

6. with the constraint that there is no recursion in the calling stack to keep the stack finite.

must be located at the end of the expansion. Thus in a PSDG, a policy might be expanded into a sequence of policies at the lower level which will be executed one after another before control is returned to the higher level policy. The implicit assumption here is that when a policy in the sequence terminates, it always does so at a state where the next policy in the sequence is applicable. Given this assumption, in the language of the AHMM we can define a *compound* policy $\pi^k$ as a policy that simply and orderly executes a sequence of policies at the lower level $\pi^{k-1}_{(1)}, \ldots, \pi^{k-1}_{(m)}$, independent of the current state. A PSDG is then equivalent to an AHMM if compound policies of this form are allowed.

Since the AMM closely follows the models used in abstract probabilistic planning, it can be used to model and recognise the behaviours of any autonomous agent whose decision making process is equivalent to an abstract MDP. It is also useful as a formal language for specifying contingent plans whose execution can then be monitored using the policy recognition algorithm. The language is also rich enough to specify a range of useful human behaviours, especially in domains where there is a natural hierarchical decomposition of the state space. Section 6 presents an application of the AHMM framework to the problem of recognising people behaviours in a complex spatial environment. Here, each policy of the AHMM represents the evolution of possible trajectories of people movement while the person performs a certain task in the environment such as heading towards a door, using the computer at a certain location, etc. The policies at different levels would represent the evolution of trajectories at different levels of abstraction. Due to the existing hierarchy in the domain, the policies can be constructed using the region-based decomposition of the state space. The environment is populated with multiple cameras divided into different zones that can provide the current location of the tracking target, albeit a noisy one. The noisy observations can be readily handled by the observation model in the AHMM. The policy recognition algorithm can then be applied to infer the person's current policy at different levels in the hierarchy.

One main restriction of the current AHMM model is that we consider only one top-level policy at a time, thus are unable to model the inter-leaving of concurrent plans. Another more subtle restriction is the assumption that a high level policy selects the lower level policies depending only on the current state. If the state space is interpreted as the states of the external environment, this assumption implies that the actor either has full observation about the current state, or at least refines its intentions based on the actor's observation about the current state only (and not the entire observation history). Note that these restrictions of the AHMM also apply in the case of the PSDG model.

## 4. Dynamic Bayesian Network Representation

In this section, we describe the Dynamic Bayesian Network (DBN) representation of the AHMM. The network serves two purposes: (1) as the tool to derive the probabilistic independence property of this stochastic model, and (2) as the computational framework for the policy recognition algorithms in Section 5.

### 4.1 Network Construction

At time $t$, let $s_t$ represent the current state, $\pi^k_t$ represent the current policy at level $k$ ($k = 0, \ldots, K$), $e^k_t$ represent the ending status of $\pi^k_t$, i.e., a boolean variable indicating whether the policy $\pi_t^k$ terminates at the current time. These variables would make up the current time-slice of the full DBN. For our convenience, the notation $\pi_t^{all}$ refers to the set of all the current policies $\{\pi_t^K, \ldots, \pi_t^0\}$. Before presenting the full network, we first describe the two sub-structures that model how policies are terminated and selected. The full DBN can then be easily constructed from these sub-structures.

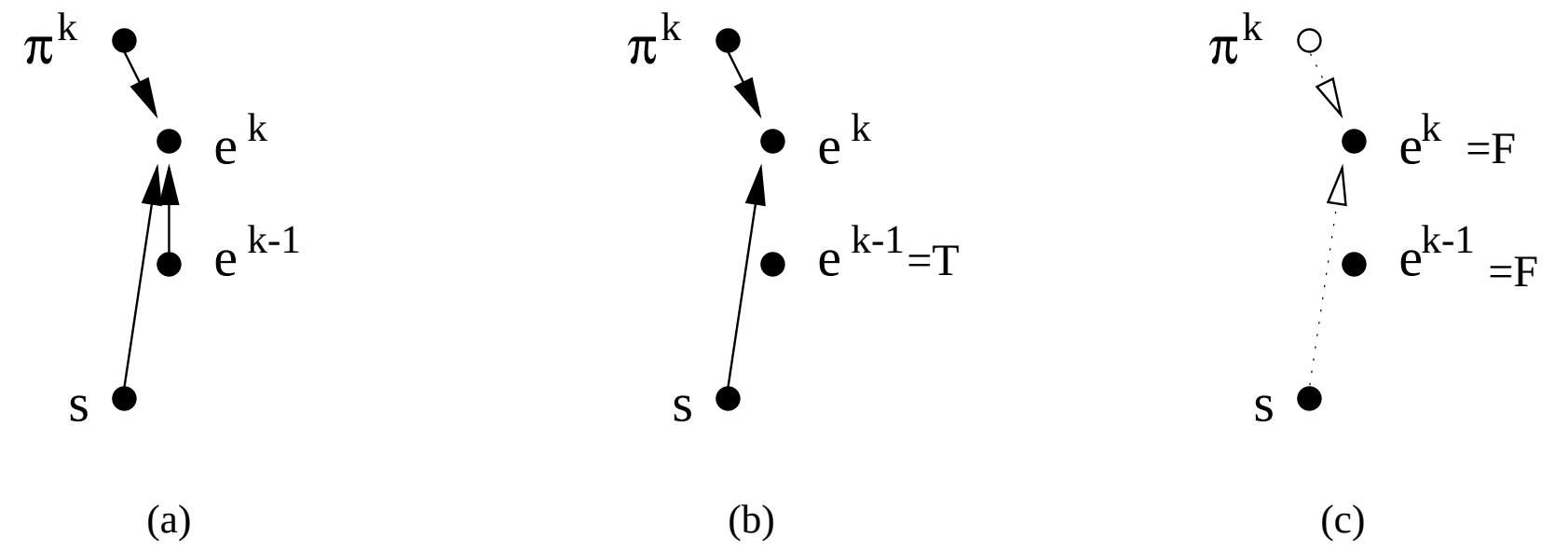

Figure 6: Sub-network for policy termination

#### 4.1.1 Policy Termination

From the definition of abstract policies, a level-$k$ policy $\pi_t^k$ terminates only if the lower level policy $\pi_t^{k-1}$ terminates, and if so, $\pi_t^k$ terminates with probability $\beta_{\pi_t^k}(s_t)$. In the Bayesian network representation, the terminating status $e_t^k$ therefore has three parent nodes: $\pi_t^k$, $s_t$, and $e_t^{k-1}$ (Fig. 6(a)).

The parent variable $e_t^{k-1}$ however plays a special role. If $e_t^{k-1} = T$, meaning the lower level policy terminates at the current time, $\Pr(e_t^k = T \mid \pi_t^k, s_t) = \beta_{\pi_t^k}(s_t)$ which gives the conditional probability of $e_t^k$ given the other two parent variables (Fig. 6(b)). However, if $e_t^{k-1} = F$, $\pi_t^k$ should not terminate and so $e_t^k = F$. Therefore, given that $e_t^{k-1} = F$, $e_t^k$ is deterministically determined and is independent of the other two parent variables $\pi_t^k$ and $s_t$. Using the notion of context-specific independence (CSI) (Boutilier et al., 1996), we can then safely remove the links from the other two parents to $e_t^k$ in the context that $e_t^{k-1}$ is false (Fig. 6(c)).

At the bottom level, since the primitive action always terminates immediately, $e_t^0 = T$ for all $t$. Since we are modelling the execution of a single top-level policy $\pi^K$, we can assume that the top-level policy does not terminate and remains unchanged: $e_t^K = F$ and $\pi_t^K = \pi^K$ for all $t$. Also, note that $e_t^l = T \Rightarrow e_t^k = T$ for all $k \leq l$, and $e_t^l = F \Rightarrow e_t^k = F$ for all $k \geq l$. Thus, at each time $t$, there exists $0 \leq l_t < K$ such that $e_t^k = T$ for all $k \leq l_t$, and $e_t^k = F$ for all $k > l_t$. The variable $l_t$ is termed the highest level of termination at time $t$. Knowing the value of $l_t$ is equivalent to knowing the terminating status of all the current policies.

#### 4.1.2 Policy Selection

The current policy $\pi_t^k$ in general is dependent on the higher level policy $\pi_t^{k+1}$, the previous state $s_{t-1}$, the previous policy at the same level $\pi_{t-1}^k$ and its ending status $e_{t-1}^k$. In the Bayesian network, $\pi_t^k$ thus has these four variables as its parents (Fig. 7(a)). This depen-dency can be further broken down into two cases, depending on the value of the parent node $e^k_{t-1}$.

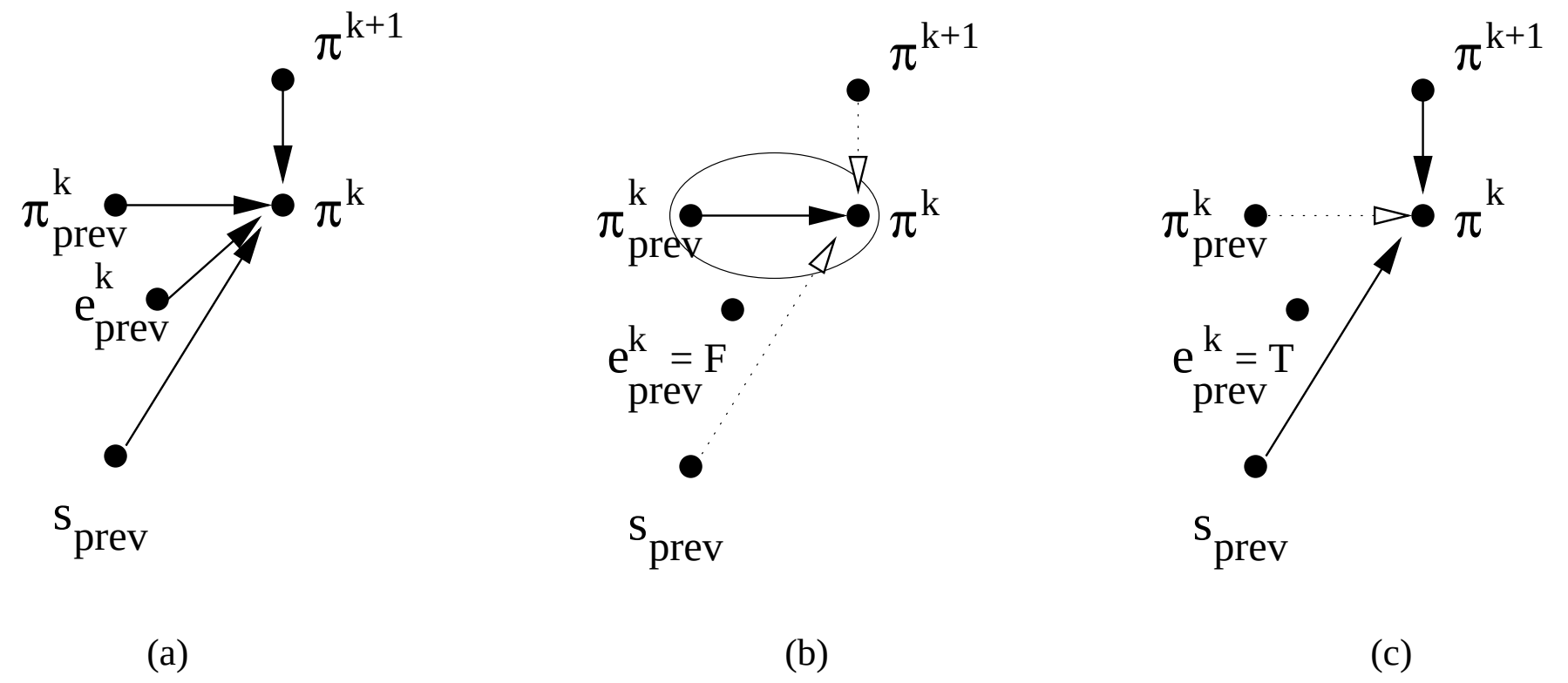

Figure 7: Sub-network for policy selection

If the previous policy has not terminated ($e^k_{t-1} = F$), the current policy is the same as the previous one: $\pi^k_t = \pi^k_{t-1}$, and the variable $\pi^k_t$ is thus independent of $\pi^{k+1}_t$ and $s_{t-1}$. Therefore, in the context $e^k_{t-1} = F$, the two links from $\pi^{k+1}_t$ and $s_{t-1}$ to the current policy can be removed, and the two nodes $\pi^k_t$ and $\pi^k_{t-1}$ can be merged together (Fig. 7(b)).

If the previous policy has terminated ($e^k_{t-1} = T$), the current policy is selected by the higher level policy with probability $\Pr(\pi^k_t \mid \pi^{k+1}_t, s_{t-1}) = \sigma_{\pi^{k+1}_t}(s_{t-1}, \pi^k_t)$. In this context, $\pi^k_t$ is independent of $\pi^k_{t-1}$ and the corresponding link in the Bayesian network can be removed (Fig. 7(c)).

#### 4.1.3 The Full DBN

The full dynamic Bayesian network can be constructed for all the policy, ending status, and state variables by putting the sub-networks for policy termination and selection together (Fig. 8). At the top level, since $e^K_t = F$, we can remove the ending status nodes and merge all the $\pi^K_t$ into a single node $\pi^K$. At the base level, since $e^0_t = T$, we can remove the ending status nodes and also the links from $\pi^0_t$ to $\pi^0_{t+1}$. To model the observation of the hidden states, an observation layer can be attached to the state layer as shown in Fig. 8.

Suppose that we are given a context where each of the variable $e^k_t$ is known. We can then modify the full DBN using the corresponding link removal and node merging rules. The result is a more intuitive tree-shaped network in Fig. 9, where all the policy nodes corresponding to the same policy for its entire duration are grouped into one. The grouping can be done since knowing the value of each $e^k_t$ is equivalent to knowing the exact duration of each policy in the hierarchy. One would expect that performing probabilistic inference on this structure is more simple than that of the full DBN in Fig. 8. In particular, if the state sequence is known, the remainder of the network in Fig. 9 becomes singly-connected, i.e., a directed graph with no undirected cycles, allowing inference to be performed with complexity linear to the size of the network (Pearl, 1988). The policy recognition algorithms that follow later exploit extensively this particular tractable case of the AHMM.

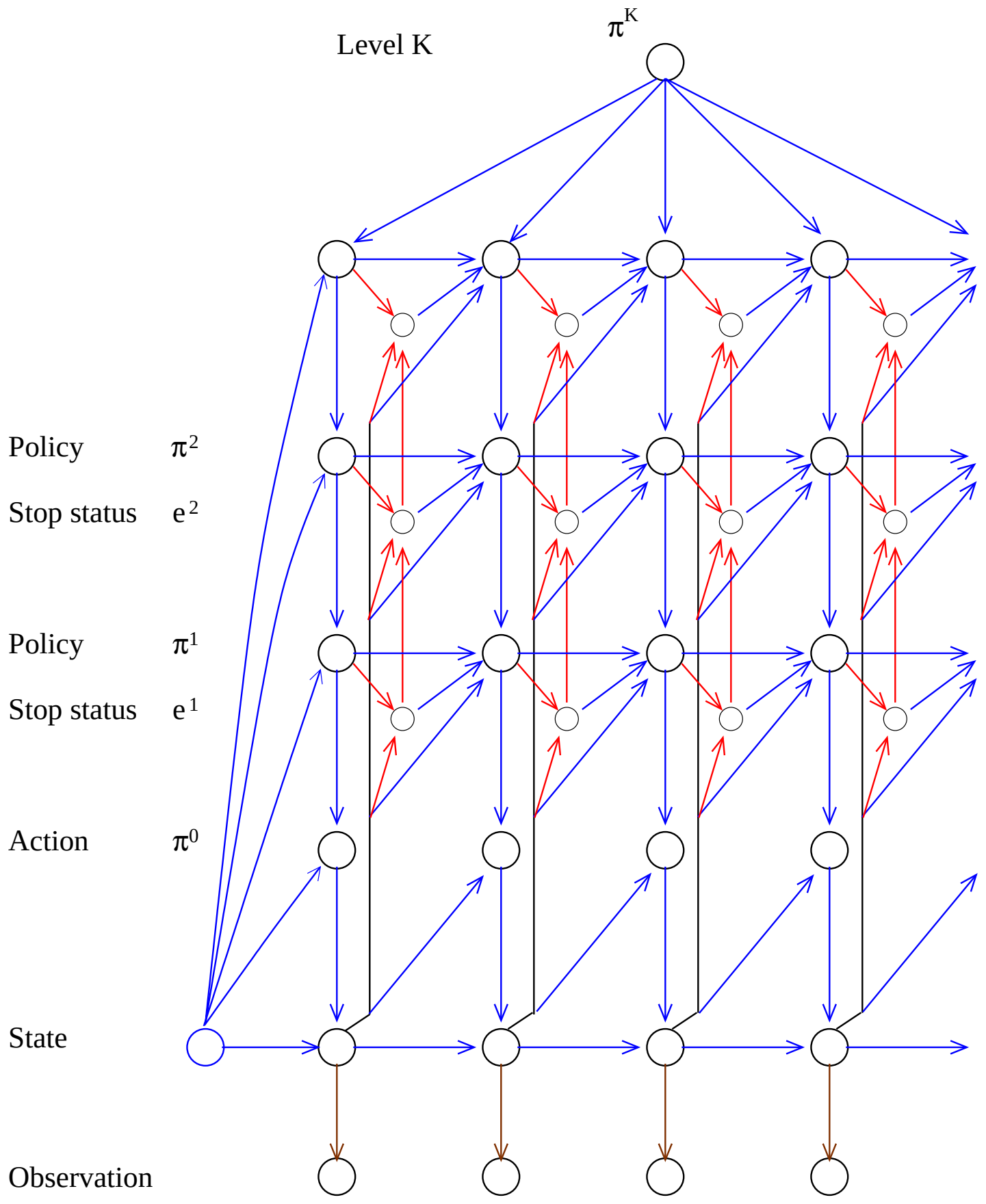

Figure 8: The DBN representation of the Abstract Hidden Markov Model

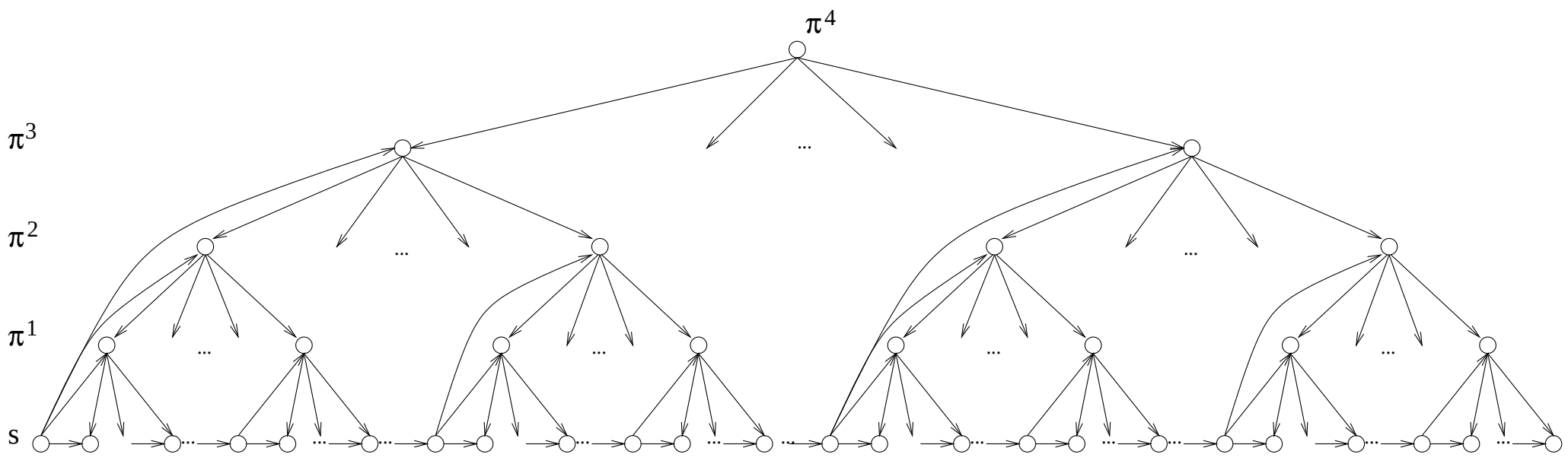

Figure 9: Simplified network if the duration of each policy is known (action nodes are omitted for clarity)

### 4.2 Conditional Independence in the Current Time-Slice

The above discussion identifies a tractable case for the AHMM, but it requires the knowledge of the entire history of the state and the policy ending status variables. In this subsection, we focus on the conditional independence property of the nodes in the current time-slice: $s_t, \pi_t^0, \ldots, \pi_t^K$. Since these nodes will make up the belief state of any future inference algorithm for our AHMM, any independence properties among these variables, if exploited, can provide a more compact representation of the belief state and reduce the inference complexity.

Due to the way policies are invoked in the AMM, we can make an intuitive remark that the higher level policies can only influence what happens at the lower level through the current level. More precisely, for a level $k$ policy $\pi_t^k$, if we know its starting state, the course of its execution is fully determined, where being determined here means without influence from what is happening at the higher levels. Furthermore, if we also know how long the policy has been executed, or equivalently its starting time, the current state of its execution is also determined. Thus, the higher level policies can only influence the current state of execution of $\pi_t^k$ either through its starting state or starting time. In other words, if we know $\pi_t^k$ together with its starting time and starting state, then the current higher level policies are completely independent of the current lower level policies and the current state. The theorem 1 below formally states this in a precise form. Note that the condition obtained is the strictest: if one of the three conditional variables is unknown, there are examples of AMMs in which the higher level policies can influence the lower level ones.

**Theorem 1.** *Let $\tau_t^k$ and $b_t^k$ be two random variables representing the starting time and the starting state, respectively, of the current level-k policy $\pi_t^k$: $\tau_t^k = \max\{t' < t \mid e_{t'}^k = T\}$ and $b_t^k = s_{\tau_t^k}$. Let $\pi_t^{>k} = \{\pi_t^{k+1}, \ldots, \pi_t^K\}$ denote the set of current policies from level $k+1$ up to $K$, and $\pi_t^{<k} = \{s_t, \pi_t^0, \ldots, \pi_t^{k-1}\}$ denote the set of current policies from level $k-1$ down to 0 together with the current state. We have:*

$$\pi_t^{>k} \perp \pi_t^{<k} \mid \pi_t^k, b_t^k, \tau_t^k \tag{7}$$

*Proof.* We sketch here an intuitive proof of this theorem through the use of the Bayesian network manipulation rules for context-specific independence which have been discussed in 4.1.1 and 4.1.2. An alternative proof that does not use CSI can be found in (Bui et al., 2000).

We first note that the theorem is not obvious by looking at the full DBN in Fig. 8. Therefore, we shall proceed by modifying the network structure in the context that we know $\tau_t^k$.

At time $\tau_t^k$, all the policies at level $k$ and below must terminate: $e_{\tau_t^k}^l = T$ for all $l \leq k$. Thus we can remove all the links from these policies to the new policies at time $\tau_t^k + 1$.

On the other hand, from time $\tau_t^k + 1$ until the current time $t$, all the policies at level $k$ and above must not terminate: $e_{t'}^l = F$ for all $l \geq k$, $\tau_t^k + 1 \leq t' < t$. Thus we can group all the policies at level $l \geq k$ between time $\tau_t^k + 1$ and $t$ into one node representing the current policy at level $l$.

These two network manipulation steps result in a network with the structure shown in Fig. 10. Once the modified network structure is obtained, we can observe that $\pi_t^{>k}$ and $\pi_t^{<k}$ are d-separated by $\pi_t^k$ and $b_t^k$ in the new structure. Thus $\pi_t^{>k}$ and $\pi_t^{<k}$ are independent given $\pi_t^k$, $b_t^k$ and $\tau_t^k$. □

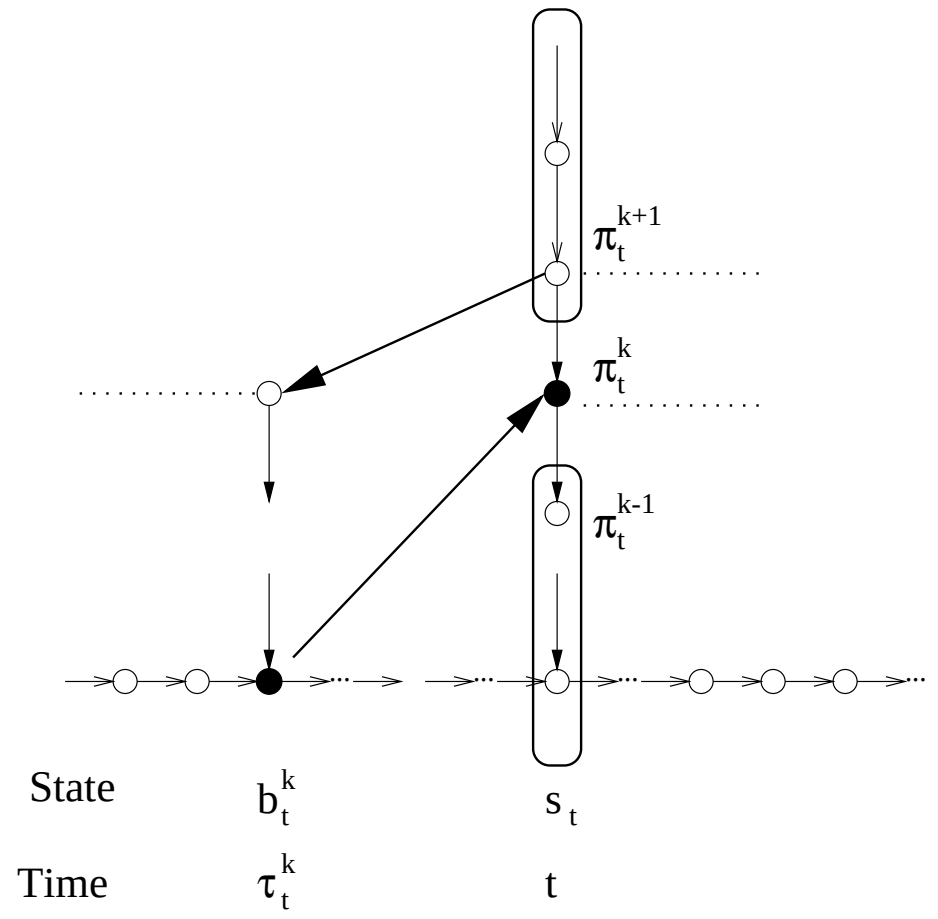

Figure 10: Network structure after being conditioned on $\tau_t^k$

## 5. Policy Recognition

In this section we begin to address the problem of policy recognition in the framework of the AHMM. We assume that a policy hierarchy is given and is modelled by an AHMM, however the top level policy and the details of its execution are unknown. The problem is then to determine the top level policy and other current policies at the lower levels given the current sequence of observations. In more concrete terms, we are interested in the conditional probability:

$$\Pr(\pi_t^K, \ldots, \pi_t^0 \mid \tilde{o}_{t-1})$$

and especially, the marginals:

$$\Pr(\pi_t^k \mid \tilde{o}_{t-1}), \text{ for all levels } k$$

Computing these probabilities gives us the information about the current policies at all levels of abstraction, from the current action ($k = 0$), to the top-level policy ($k = K$), taking into account all the observations that we have up to date.

In typical monitoring situations, these probabilities need to be computed "online", as each new observation becomes available. To do this, it is required to update the belief state (filtering distribution) of the AHMM at each time point $t$. This problem is generally intractable unless the belief state has an efficient representation that affords a closed form update procedure. In our case, the belief state is a joint distribution of $K + 3$ discrete variables: $\Pr(\pi_t^K, \ldots, \pi_t^0, s_t, l_t \mid \tilde{o}_t)$. Without any further structure imposed on the belief state, the complexity for updating it is exponential in $K$.

To cope with this complexity, one generally has to resort to some form of approximation to trade off accuracy for computational resources. On the other hand, the analysis of the AHMM network in the previous section suggests that the problem of inference in the AHMM can be tractable in the special case when the history of the state and terminating status variables is known. Motivated by this property of the AHMM, our main aim in this section is to derive a hybrid inference scheme that combines both approximation and tractable exact inference for efficiency. We first treat the special case of policy recognition where the belief state of the AHMM has a tractable structure in 5.1. We then present a hybrid inference scheme for the general case using the Rao-Blackwellised Sequential Importance Sampling (RB-SIS) method in 5.2.

### 5.1 Policy Recognition: the Tractable Case

Here, we address the policy recognition problem under two assumptions: (1) the state sequence can be observed with certainty, and (2) the exact time when each policy starts and ends is known. More precisely, our observation at time $t$ includes the state history $\tilde{s}_t = (s_0, \ldots, s_t)$ and the policy termination history $\tilde{l}_t = (l_0, \ldots, l_t)$. The belief state that we need to compute in this case is $\mathcal{B}_t = \Pr(\pi_t^{all}, s_t, l_t \mid \tilde{s}_{t-1}, \tilde{l}_{t-1})$ and its posterior after absorbing the observation at time $t$: $\mathcal{B}_{t+} = \Pr(\pi_t^{all} \mid \tilde{s}_t, \tilde{l}_t)$.

The first assumption means that the observer always knows the true current state and is often referred to as "full observability". When the states are fully observable, we can ignore the observation layer $\{o_t\}$ in the AHMM and thus only have to deal with the AMM instead. The second assumption means that the observer is fully aware when the current policy ends and a new policy begins. If the policy hierarchy is constructed from the region-based decomposition of the state space (subsection 3.2), the termination status can be inferred directly from the state sequence. Thus for SRD policy hierarchies, only the full observability condition is needed since the second assumption is subsumed by the first and can be left out. Except for SRD policy hierarchies, these two assumptions are usually too restrictive for the policy recognition algorithm presented here to be useful by itself. However, the algorithm for this special case will form the exact step in the hybrid algorithm presented in subsection 5.2 for the general case.

#### 5.1.1 Representation of the belief state

We first look at the conditional joint distribution $\Pr(\pi_t^{all}, s_t \mid \tilde{s}_{t-1}, \tilde{l}_{t-1})$. From the termination history $\tilde{l}_{t-1}$, we can derive precisely the starting time of the current level-k policy:

$$\tau_t^k = \max\{0\} \cup \{t' < t \mid e_{t'}^k = T\} = \max\{0\} \cup \{t' < t \mid l_{t'} \geq k\}$$

On the other hand, knowing the starting time together with the state history also gives us the starting state $b_t^k$. Thus, both the starting time and the starting state of $\pi_t^k$ can be derived from $\tilde{s}_{t-1}$ and $\tilde{l}_{t-1}$. From Theorem 1, we obtain for all level $k$:

$$\pi_t^{>k} \perp \pi_t^{<k} \mid \pi_t^k, \tilde{s}_{t-1}, \tilde{l}_{t-1}$$

In other words, given $\tilde{s}_{t-1}$ and $\tilde{l}_{t-1}$, the conditional joint distribution of $\{\pi_t^K, \ldots, \pi_t^0, s_t\}$ can be represented by a Bayesian network with a simple chain structure. We denote this chain network by $\mathcal{C}_t \equiv \Pr(\pi_t^{all}, s_t \mid \tilde{s}_{t-1}, \tilde{l}_{t-1})$ and term it the *belief chain* for the role it plays in the representation of the belief state (Fig. 11(a)). If a chain is drawn so that all links point away from the level-k node, we say that the chain has root at level $k$. The root of the chain can be moved from $k$ to another level $k'$ simply by reversing the links lying between $k$ and $k'$ using the standard link-reversal operation for Bayesian networks (Shachter, 1986).

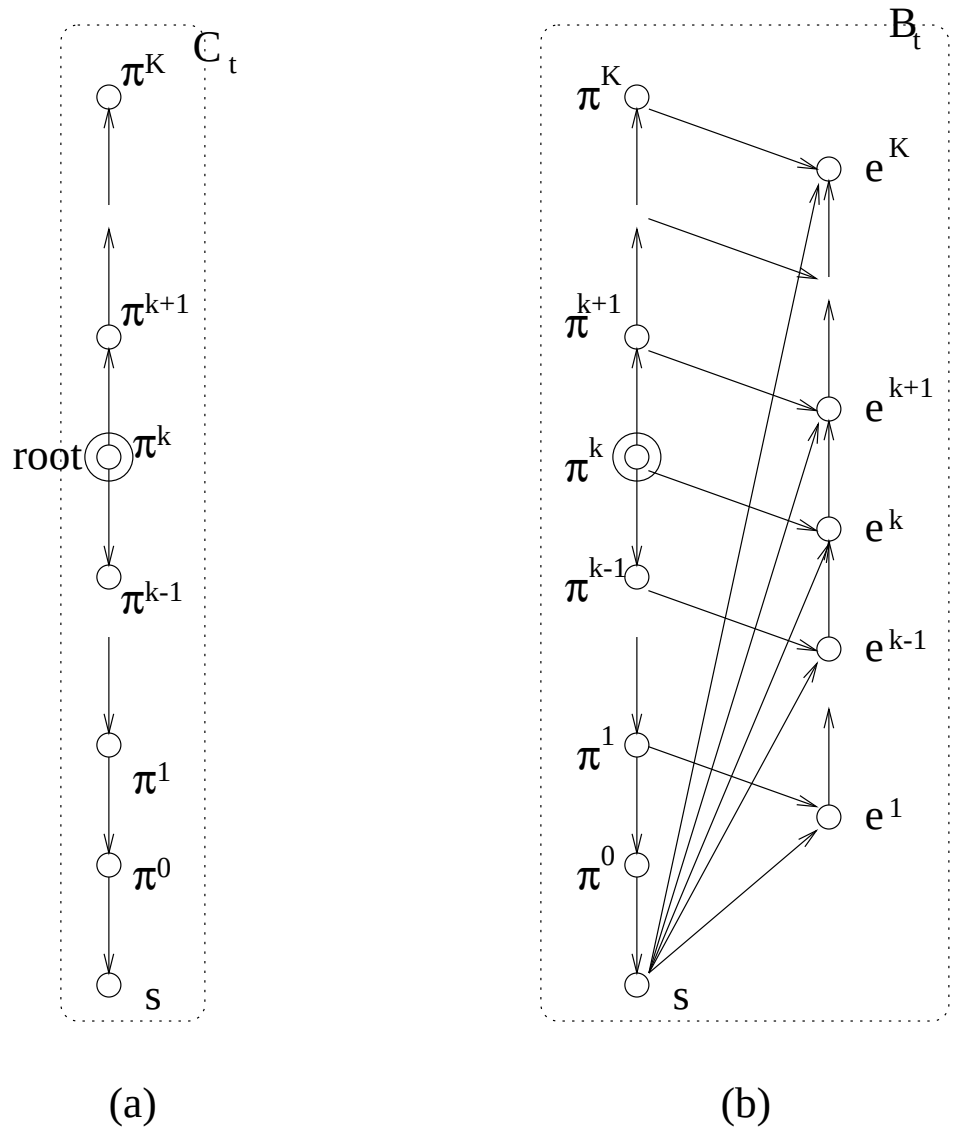

Figure 11: Representation of the belief state

Each node in the belief chain also has a manageable size. In principle, the domain of $\pi_t^k$ is $\Pi^k$, the set of all policies at level $k$, and the domain of $s_t$ is $S$, the set of all possible states. When $K$ is large, we basically want to model a larger state space, and the set of policies to cover this state space is also large. The sizes of these domains would most likely grow exponential w.r.t. $K$. However, given a particular state, the number of policies applicable at that state would remain relatively constant and independent of $K$. For each policy $\pi_t^k$, we know its starting state $b_t^k$, which implies that $\pi_t^k \in \Pi^k(b_t^k)$, the set of all level-$k$ policies applicable at $b_t^k$. Thus $\Pi^k(b_t^k)$ can be used as the "local" domain for $\pi_t^k$ to avoid the exponential dependency on $K$. Similarly, the domain for $s_t$ can be taken as the set of neighbouring states of $s_{t-1}$ (reachable from $s_{t-1}$ by performing one primitive action). For a given state, we term the maximum number of relevant objects (applicable policies/actions, neighbouring states) at a single level the *degree of connectivity* $\mathcal{N}$ of the domain being modelled. The size of the conditional probability table for each link of the belief chain is then $O(\mathcal{N}^2)$, and the overall size of the belief chain is $O(K\mathcal{N}^2)$.

We now can construct the belief state $\mathcal{B}_t$ from $\mathcal{C}_t$. Since the current terminating status is solely determined by the current policies and the current state, the belief state $\mathcal{B}_t$ can be factorised into:

$$\Pr(\pi_t^{all}, s_t, l_t \mid \tilde{s}_{t-1}, \tilde{l}_{t-1}) = \Pr(l_t \mid \pi_t^{all}, s_t) \Pr(\pi_t^{all}, s_t \mid \tilde{s}_{t-1}, \tilde{l}_{t-1}) = \Pr(l_t \mid \pi_t^{all}, s_t)\mathcal{C}_t$$

Note that the variable $l_t$ is equivalent to the set of variables $\{e_t^K, \ldots, e_t^1\}$. Thus, the full belief state $\mathcal{B}_t$ can be realised by adding to $\mathcal{C}_t$ the links from the current policies and the current state to the terminating status variables $e_t^k$ (Fig. 11(b)). The size of the belief state would still be $O(K\mathcal{N}^2)$. If the state is a composite of many orthogonal variables, a factored representation can be used so that the size of the belief state representation does not depend exponentially on the dimensionality of the state space. We discuss factored representations further under subsection 5.2.2.

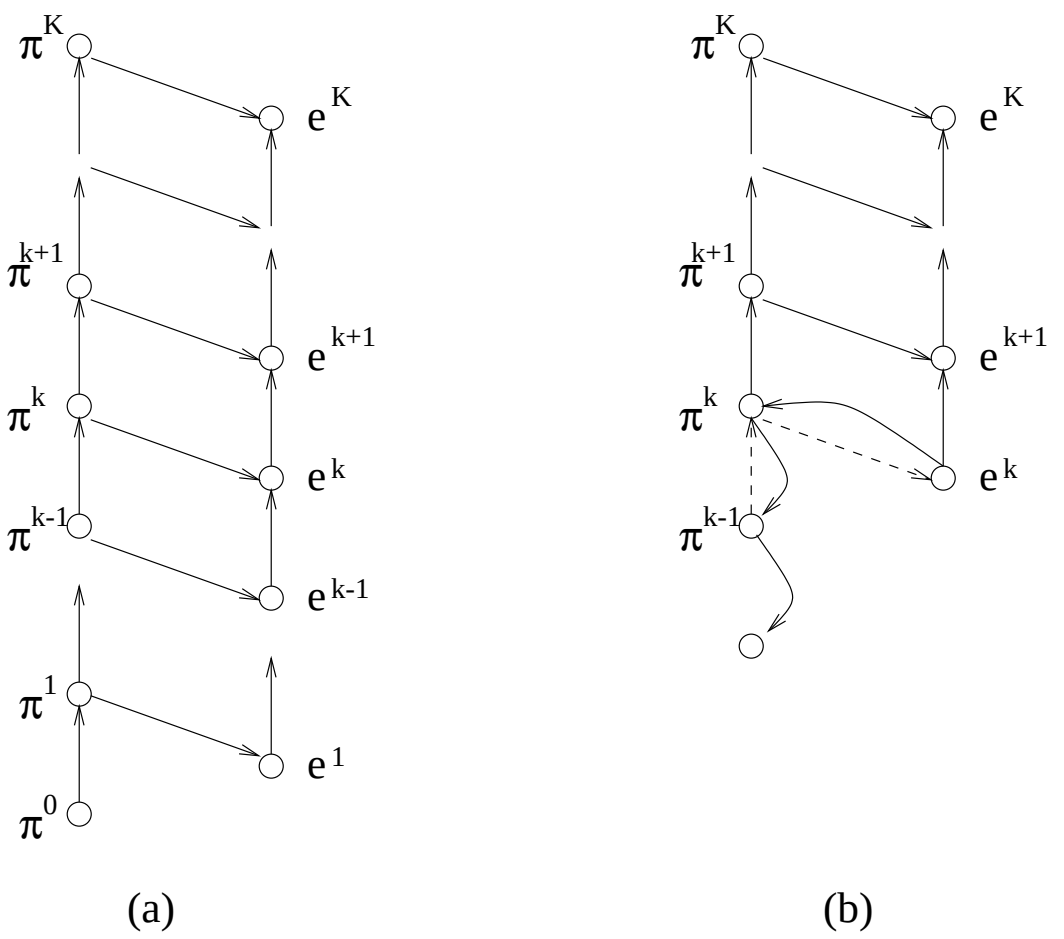

Figure 12: Belief state updating: from $\mathcal{B}_t$ to $\mathcal{B}_{t+}$

#### 5.1.2 Updating the belief state

Since the belief state $\mathcal{B}_t$ can be represented by a simple belief network in Fig. 11(b), we can expect that a general exact inference method for updating the belief state such as (Kjærulff, 1995) will work efficiently. However, this general method works with undirected network representation of the belief state distribution which can be inconvenient for us later on when we want to sample from such a distribution. Here, we describe an algorithm that updates the belief state in the closed form given by the directed network in Fig. 11(b).

Assuming that we have a complete specification of the belief state $\mathcal{B}_t$, i.e., all the parameters for its Bayesian network representation, we need to compute the parameters for the new network $\mathcal{B}_{t+1}$. This is done in two steps, as in the standard "roll-over" of the belief state of a DBN: (1) absorbing the new evidence $s_t$, $l_t$ and (2) projecting the belief state into the next time step.

The first step corresponds to the instantiation of the variables $s_t$, $e_t^1, \ldots, e_t^K$ in the Bayesian network $\mathcal{B}_t$ to obtain $\mathcal{B}_{t+}$ which is the conditional joint distribution of $\pi_t^K, \ldots, \pi_t^0$. By checking the conditional independence relationships in Fig. 11(b), it is easy to see that $\mathcal{B}_{t+}$ again has a simple chain network structure. Thus, conceptually, the problem here is to update the parameters of the chain $\mathcal{C}_t$ so as to absorb the given evidence to form a new chain $\mathcal{B}_{t+}$. This can be done by a number of link-reversal steps as follows.

To instantiate $s_t$, we first move the root of the chain $\mathcal{C}_t$ to $s_t$. The variable $s_t$ then has no parents and can be instantiated and deleted from the network (Fig. 12(a)).

To instantiate $l_t$ which is equivalent to the value assignment ($e_t^K = F, \ldots, e_t^{l_t+1} = F, e_t^{l_t} = T, \ldots e_t^1 = T$), starting from $k = 1$, we iteratively reverse the links from $\pi_t^{k-1}$ to $\pi_t^k$ and from $\pi_t^k$ to $e_t^k$ (Fig. 12(b)). In algebraic forms, the first link reversal operation corresponds to computing the following probabilities:

$$\Pr(\pi_t^k \,|\, s_t, e_t^1, \ldots, e_t^{k-1}) = \sum_{\pi_t^{k-1}} \Pr(\pi_t^k \,|\, \pi_t^{k-1}) \Pr(\pi_t^{k-1} \,|\, s_t, e_t^1, \ldots, e_t^{k-1}) \tag{8}$$

$$\Pr(\pi_t^{k-1} \,|\, \pi_t^k, s_t, e_t^1, \ldots, e_t^{k-1}) \propto \Pr(\pi_t^k \,|\, \pi_t^{k-1}) \Pr(\pi_t^{k-1} \,|\, s_t, e_t^1, \ldots, e_t^{k-1}) \tag{9}$$

and the second link reversal corresponds to:

$$\Pr(\pi_t^k \,|\, s_t, e_t^1, \ldots, e_t^k) \propto \Pr(e_t^k \,|\, \pi_t^k, s_t, e_t^{k-1}) \Pr(\pi_t^k \,|\, s_t, e_t^1, \ldots, e_t^{k-1}) \tag{10}$$

Effectively, the $k$-th link reversal step positions the root of the chain $\mathcal{C}_t$ at $\pi_t^k$ and absorbs the evidence $e_t^k$. By repeating this link reversal operations with $k = 1, \ldots, l_t + 1$, we obtain a new chain for $\mathcal{B}_{t+}$ which has root at level $l_t + 1$. Note that there is no need to incorporate the instantiations $e_t^k = F$ for $k > l_t + 1$ since they are the direct consequences of the instantiation $e_t^{l_t+1} = F$. The parameters of the chain $\mathcal{B}_{t+}$ are given below. The upward links remain the same as those of $\mathcal{C}_t$, while the marginal at level $l_t + 1$ and the downward links are obtained as the results of the link reversal operations above:

$$\begin{aligned}
\Pr(\pi_t^{k+1} \,|\, \pi_t^k, s_t, l_t) &= \Pr(\pi_t^{k+1} \,|\, \pi_t^k), k \geq l_t + 1 \\
\Pr(\pi_t^k \,|\, s_t, l_t) &= \Pr(\pi_t^k \,|\, s_t, e_t^1, \ldots, e_t^k), k = l_t + 1 \\
\Pr(\pi_t^{k-1} \,|\, \pi_t^k, s_t, l_t) &= \Pr(\pi_t^{k-1} \,|\, \pi_t^k, s_t, e_t^1, \ldots, e_t^{k-1}), k \leq l_t
\end{aligned}$$

In the second step, we continue to compute $\mathcal{C}_{t+1}$ from $\mathcal{B}_{t+}$. Since all the policies at levels higher than $l_t$ do not terminate, $\pi_{t+1}^{>l_t} = \pi_t^{>l_t}$, and we can retain this upper sub-chain from $\mathcal{B}_{t+}$ to $\mathcal{C}_{t+1}$. In the lower part, for $k \leq l_t$, a new policy $\pi_{t+1}^k$ is created by the policy $\pi_{t+1}^{k+1}$ at the state $s_t$, and thus a new sub-chain can be formed among the variables $\pi_{t+1}^{<l_t}$ with parameters $\Pr(\pi_{t+1}^k \,|\, \pi_{t+1}^{k+1}, s_t) = \sigma_{\pi_{t+1}^{k+1}}(s_t, \pi_{t+1}^k)$. Note that the domain of the newly-created node $\pi_{t+1}^k$ is $\Pi^k(s_t)$. The new chain $\mathcal{C}_{t+1}$ is then the combination of these two sub-chains, which will be a chain with root at level $l_t + 1$ (see Fig. 13). Once we have the chain $\mathcal{C}_{t+1}$, the new belief state $\mathcal{B}_{t+1}$ can be obtained by simply adding the terminating status variables $\{e_{t+1}^k\}$ to $\mathcal{C}_{t+1}$.

This completes the procedure for updating the belief state from $\mathcal{B}_t$ to $\mathcal{B}_{t+1}$, thus allowing us to compute the belief state $\mathcal{B}_t$ at each time step. Although the belief state is the joint distribution of all the current variables, due to its simple structure, the marginal distribution of a single variable can be computed easily. For example, if we are only interested in the current level-$k$ policy $\pi_t^k$, the marginal probability $\Pr(\pi_t^k \,|\, \tilde{s}_{t-1}, \tilde{l}_{t-1})$ is simply the marginal at the level-$k$ node in the chain $\mathcal{C}_t$, and can be readily obtained from the chain parameters.

The complexity of the belief state updating procedure at time $t$ is proportional to $l_t$ since it only needs to modify the bottom $l_t$ levels of the belief state. On the other hand, the probability that the current policy at level $l$ terminates can be assumed to be exponentially small w.r.t. $l$. Thus, the average updating complexity at each time-step is $O(\sum_l l/exp(l))$ which is constant-bounded, and thus does not depend on the number of levels in the policy hierarchy. In terms of the number of policies and states, the updating complexity is linear to the size of a policy node in the belief chain, thus is linear to the degree of connectivity of the domain.

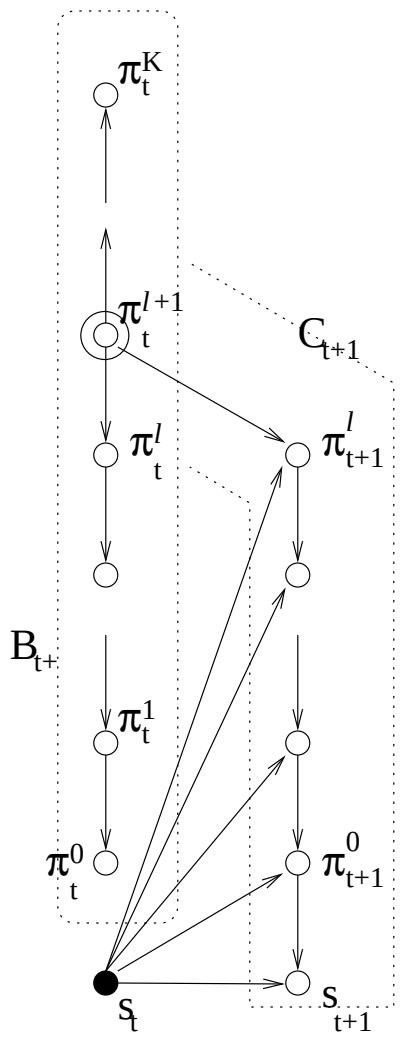

Figure 13: Belief state updating: from $\mathcal{B}_{t+}$ to $\mathcal{C}_{t+1}$

### 5.2 Policy Recognition: The General Case

We now return to the general case of policy recognition, i.e., without the two assumptions of the previous subsection. This makes the inference tasks in the AHMM much more difficult. Since neither the starting times nor the starting states of the current policies are known with certainty, theorem 1 cannot be used. Thus, the set of current policies no longer forms a chain structure as it did in $\mathcal{C}_t$ since the conditional independence properties of the current time-slice no longer hold. We therefore cannot hope to represent the belief state by a simple structure as we did previously. An exact method for updating the belief state will thus have to operate on a structure with size exponential in $K$, and is bound to be intractable when $K$ is large.

To cope with this complexity, an approximation scheme such as sequential importance sampling (SIS) (Doucet et al., 2000b; Liu & Chen, 1998; Kanazawa et al., 1995) can be employed. In our previous work (Bui, Venkatesh, & West, 1999), we have applied an SIS method known as the likelihood weighting with evidence reversal (LW-ER) (Kanazawa et al., 1995) to an AHMM-like network structure. However the SIS method needs to sample in the product space of all the layers of the AHMM and thus becomes less accurate and inefficient with large $K$. The key to get around this inefficiency is to utilise the special structure of the AHMM, particularly, its special tractable case, to keep the set of variables that need to be sampled to a minimum.

The improvement of the SIS method to achieve this is has been presented in subsection 2.5 in the name of the Rao-Blackwellised SIS (RB-SIS) method. Rao-Blackwellisation specifically allows the marginalisation of some variables analytically and only samples the remaining variables. As a result, this reduces the averaged error, measured as the variance of the estimator (Casella & Robert, 1996).

In order to apply RB-SIS to the AHMM, the main problem is to identify which variables should be used as the Rao-Blackwellising variables and should still be sampled, with the remaining variables being marginalised analytically. The key to choosing the Rao-Blackwellising variables, as we have shown in 2.5, is so that if those variables can be observed, the Rao-Blackwellised belief state becomes tractable. In subsection 5.1, we have demonstrated that if the state history $\tilde{s}_t$ and the terminating status history $\tilde{l}_t$ can be observed then the belief state has a simple network structure and can be updated with constant average complexity. Thus, $(s_t, l_t)$ can be used conveniently as the Rao-Blackwellising variable $r_t$. Note that the variables $\tilde{l}_t$ are the context variables which help to simplify the network structure of the AHMM, while the state variables $\tilde{s}_t$ help to make the remaining network singly-connected so that exact inference can operate efficiently (see subsection 4.1.3).

### 5.2.1 RB-SIS for AHMM

We now discuss the specific application of RB-SIS to the problem of belief state updating and policy recognition in the AHMM. Our main objective is to use RB-SIS to estimate the conditional probability of the policy currently being executed at level-$k$ given the current sequence of observations $\Pr(\pi^k_{t+1} \mid \tilde{o}_t)$.

Mapping the RB-SIS general framework in subsection 2.5 to the AHMM structure, the set of all current variables $x_t$ is now the set of current policies, terminating status nodes, and the current state: $x_t = (\pi^{all}_t, s_t, l_t)$. The probability under estimation $\Pr(\pi^k_{t+1} \mid \tilde{o}_t)$ can be viewed as an expectation by letting $f(\pi^{all}_t, s_t, l_t) = \Pr(\pi^k_{t+1} | \pi^{all}_t, s_t, l_t)$ so that:

$$\bar{f} = \sum_{\pi^{all}_t, s_t, l_t} \Pr(\pi^k_{t+1} | \pi^{all}_t, s_t, l_t) \Pr(\pi^{all}_t, s_t, l_t | \tilde{o}_t) = \Pr(\pi^k_{t+1} \mid \tilde{o}_t)$$

Using RB-SIS to estimate this expectation, we shall split $x_t$ into two sets of variables: the set of RB variables $r_t = (s_t, l_t)$, and the set of remaining variables $z_t = \pi^{all}_t$ which is the set of all the current policies. The functional $h$, which depends only on the RB variables and is obtained from $f$ by integrating out the remaining variables (Eq. (5)), now has the form:

$$h(\tilde{r}_t) = h(\tilde{s}_t, \tilde{l}_t) = \sum_{\pi^{all}_t} \Pr(\pi^k_{t+1} \mid \pi^{all}_t, s_t, l_t) \Pr(\pi^{all}_t \mid \tilde{s}_t, \tilde{l}_t, \tilde{o}_t) = \Pr(\pi^k_{t+1} \mid \tilde{s}_t, \tilde{l}_t) \tag{11}$$

which is the marginal $\mathcal{C}_{t+1}(\pi^k_{t+1})$ from the belief chain at time $t+1$.

The RB belief state, which is the belief state of the AHMM when the RB variables are known, becomes:

$$\mathcal{R}_t = \Pr(\pi^{all}_t, s_t, l_t, o_t \mid \tilde{s}_{t-1}, \tilde{l}_{t-1}, \tilde{o}_{t-1}) = \Pr(\pi^{all}_t, s_t, l_t, o_t \mid \tilde{s}_{t-1}, \tilde{l}_{t-1}) \tag{12}$$

and is identical to the special belief state $\mathcal{B}_t$ discussed in subsection 5.1, except a minor modification to attach the observation variable $o_t$.

From (11) and (12), both the $h$ function and the RB belief state can be computed very efficiently using the exact inference techniques described in 5.1. Thus RB-SIS can be implemented efficiently with minimal overhead in exact inference.

The main RB-SIS algorithm for the AHMM is given in Fig. 14. Note that we only need to sample the RB variables $\tilde{s}_t$ and $\tilde{l}_t$. For each sample $i$, in addition to the weights $w^{(i)}$,

Begin
For $t = 0, 1, \ldots$
    For each sample $i = 1, \ldots, N$
        Sample $s_t^{(i)}, l_t^{(i)}$ from $\mathcal{B}_t^{(i)}(s_t, l_t |\, o_t)$
        Update weight $w^{(i)} = w^{(i)} \mathcal{B}_t^{(i)}(o_t)$
        Compute the posterior RB bel state $\mathcal{B}_{t+}^{(i)} = \mathcal{B}_t^{(i)}(\pi_t^{all} | s_t^{(i)}, l_t^{(i)}, o_t)$
        Compute the belief chain $\mathcal{C}_{t+1}^{(i)}$ from $\mathcal{B}_{t+}^{(i)}$
        Compute the new belief state $\mathcal{B}_{t+1}^{(i)}$ from $\mathcal{C}_{t+1}^{(i)}$
        Compute $h^{(i)} = \mathcal{C}_{t+1}^{(i)}(\pi_{t+1}^k)$
    Compute the estimator $\Pr(\pi_{t+1}^k \,|\, \tilde{o}_t) \approx \hat{f}_{RBSIS} = \sum_{i=1}^N h^{(i)} \tilde{w}^{(i)}$
End

Figure 14: RB-SIS for policy recognition

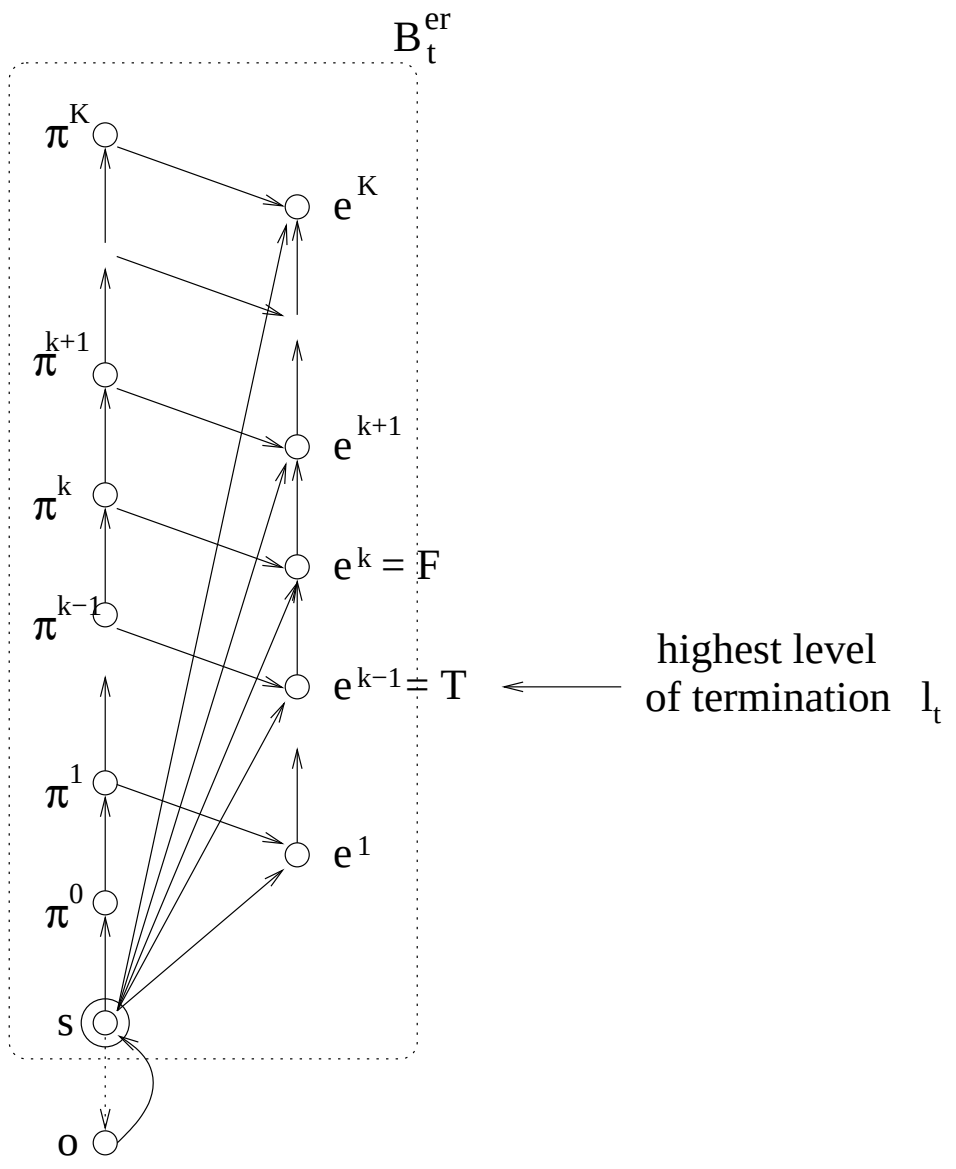

Figure 15: Sampling the Rao-Blackwellising variables in AHMM

we also maintain a parametric representation of the Rao-Blackwellised belief state $\mathcal{B}_t^{(i)}$, and the value of the $h$ function for that sample $h^{(i)}$. The weights of the samples, together with the values of the $h$ function can then be combined to yield an approximation for $\bar{f}$.

Some details on how we can obtain the new samples at each time step are worth noting here. Since we are using the optimal sampling distribution $q_t = \mathcal{B}_t(s_t, l_t |\, o_t)$ to sample the RB variables $s_t$ and $l_t$, we need to perform the evidence reversal step.[7] This can be done by positioning the root of the belief chain $\mathcal{C}_t$ at $s_t$ and reverse the link from $s_t$ to $o_t$. This gives us the network structure for $\mathcal{B}_t^{er} = \mathcal{B}_t(s_t, l_t, \pi_t^{all} | o_t)$ which is exactly the same as $\mathcal{B}_t$ (see Fig. 15), except that the evidence $o_t$ has been absorbed into the marginal distribution of $s_t$. The weight $w_t = \mathcal{B}_t(o_t)$ can also be obtained as a by-product of this evidence reversal step. In order to sample $s_t$ and $l_t$ from $\mathcal{B}_t^{er}$ without the need to compute the marginal distribution for these two variables, we can use forward sampling to sample every variable of $\mathcal{B}_t^{er}$, starting from the root node $s_t$ and proceeding upward. Since $l_t$ by definition is the highest level of policy termination, the sampling can stop at the first level $k$ where $e_t^k = F$. We can then assign $l_t$ the value $k-1$. Any unnecessary samples for the policy nodes along the way are discarded. Once we have the new samples for $s_t$ and $l_t$, the updating of the RB belief state from $\mathcal{B}_t$ to $\mathcal{B}_{t+1}$ is identical to the belief state updating procedure described in 5.1. The $h$ function can then be obtained by computing the corresponding marginal of the new belief chain $\mathcal{C}_{t+1}$.

At each time step, the complexity of maintaining a sample (sampling the new RB variables and updating the RB belief state) is again $O(l_t)$, and thus, on average, bounded by a constant. The overall complexity of maintaining every sample is thus $O(N)$ on average. If a prediction is needed, for each sample, we have to compute $h$ by manipulating the chain $\mathcal{C}_{t+1}$ with the complexity $O(K)$. Thus the complexity at the time step when a prediction needs to be made is $O(NK)$.

In comparison with the use of an SIS method such as LW-ER, the RB-SIS has the same order of computational complexity (the SIS also has complexity $O(NK)$). However, while the SIS method needs to sample every layers of the AHMM, the RB-SIS method only needs to sample two sequences of variables $\tilde{s}_t$, $\tilde{l}_t$, and avoids having to sample the $K$ policy sequences $\{\tilde{\pi}_t^k\}$. After Rao-Blackwellisation, the dimension of the sample space becomes much smaller, and more importantly, does not grow with $K$. As a result, the accuracy of the approximation by the RB-SIS method does not depend on the height of the hierarchy $K$. In contrast, due to the problems of sampling in high dimensional space, the accuracy of SIS methods tends to degrade, especially when $K$ is large.

#### 5.2.2 Performing Evidence Reversal with a Factored State Space

In many cases, the state space $S$ is the Cartesian product of many state variables representing relatively independent properties of a state: $s_t = (s_t^1, s_t^2, \ldots, s_t^M)$. Since the overall state space is very large, specifying an action by the usual transition probability matrix is problematic. It is advantageous in this case to represent the state information in a factored form, i.e., representing each state variable $s_t^m$ in a separate node rather than lumping them into a single node $s_t$. It has been shown that using factored representations, we can specify the transition probability of each action in a compact form since an action is likely to affect only a small number of state variables and the specification of the effects of actions has many regularities (Boutilier et al., 2000).

7. The term *evidence reversal* is used in this paper to refer to a general procedure in which the link to the observation node is reversed prior to sampling (Kanazawa et al., 1995), thus allowing us to sample according to the optimal sampling distribution $q_t$.

The representation of the belief chain $\mathcal{C}_t$ and also the RB belief state $\mathcal{B}_t$ can take direct advantage of this factored representation of actions. Indeed, the chain parameter $\mathcal{C}_t(s_t|\pi_t^0)$ of the link from $\pi_t^0$ to $s_t$ is precisely the transition probability for the action $\pi_t^0$ at the previous state $s_{t-1}$ (note that $s_{t-1}$ is known due to Rao-Blackwellisation). This conditional distribution can be extracted from the compact factored representation of $\pi_t^0$ in the general form of a Bayesian network of the variables $\{s_t^1, s_t^2, \ldots, s_t^M\}$. For our convenience, let us denote this Bayesian network by $\mathcal{F}(.|\pi_t^0)$. This network is usually sparse enough so that exact inference can operate efficiently. For example, in the special case where $\{s_t^1, s_t^2, \ldots, s_t^M\}$ are independent given $\pi_t^0$ and $s_{t-1}$, $\mathcal{F}$ will be factored completely into the product of $M$ marginals of $s_t^m$.

Although factored representations can be used as part of the RB belief state, care must be taken when performing evidence reversal, i.e. to reverse the link from the state variable to the observation node. In the procedure for evidence reversal discussed previously (see Fig. 15), we first position the root of $\mathcal{C}_t$ at the node $s_t$, thus need to compute and represent the distribution $\Pr(s_t)$. In the factored state space case, this becomes a joint distribution of all the state variables $\{s_t^1, s_t^2, \ldots, s_t^M\}$. Without conditioning on the current action $\pi_t^0$, the factored representation of the state variables $\{s_t^m\}$ cannot be utilised, thus resulting in complexity exponential in $M$.

The key to get around this difficulty is to always keep the specification of the distribution of the current state conditioned on the current action, not vice versa. Thus, when computing $\mathcal{B}_t^{er} = \mathcal{B}_t(.|o_t)$, we first position the root of the chain $\mathcal{C}_t$ at $\pi_t^0$, and then reverse the evidence from $o_t$ to both $\pi_t^0$ and $s_t$. In algebraic form, we use the following factorisation of the joint distribution of the current action and state given the current observation:

$$\Pr(\pi_t^0, s_t|o_t) = \Pr(s_t|\pi_t^0, o_t)\Pr(\pi_t^0|o_t) \tag{13}$$

Fig. 16 illustrates this evidence reversal procedure. In the model depicted here, $\mathcal{F}$ can be an arbitrary Bayesian network. The observation model can be specified by attaching the observation nodes $\{o_t^1, o_t^2, \ldots\}$ to the state variables. The overall network representing the distribution $\Pr(s_t, o_t \mid \pi_t^0)$ will be denoted by $\mathcal{F}^{obs}(.|\pi_t^0)$.

We first look at the first term in the RHS of (13). Let $\mathcal{F}^{er}(.|\pi_t^0, o_t)$ represent the distribution $\Pr(s_t \mid \pi_t^0, o_t)$. Note that $\mathcal{F}^{er}$ can be obtained by conditioning $\mathcal{F}^{obs}(.|\pi_t^0)$ on the observation $o_t$. This can be achieved by applying an exact inference method such as the clustering algorithm (Lauritzen & Spiegelhalter, 1988) on the network $\mathcal{F}^{obs}(.|\pi_t^0)$.

For the second term in the RHS of (13), we note that:

$$\Pr(o_t \mid \pi_t^0) = \sum_{s_t} \Pr(s_t, o_t \mid \pi_t^0) = \sum_{s_t} \mathcal{F}^{obs}(s_t, o_t|\pi_t^0)$$

This integration can be readily obtained as a by-product when performing the above clustering algorithm on $\mathcal{F}^{obs}(.|\pi_t^0)$. Once $\Pr(o_t \mid \pi_t^0)$ is known, we can compute $\Pr(\pi_t^0 \mid o_t)$ by:

$$\Pr(\pi_t^0 \mid o_t) \propto \Pr(o_t \mid \pi_t^0)\Pr(\pi_t^0)$$

This shows that the belief state after evidence reversal $\mathcal{B}_t^{er} = \mathcal{B}_t(.\mid o_t)$ still has a simple structure that exploits the independence relationships between the state variables $\{s_t^m\}$ given the current action $\pi_t^0$. Sampling the RB variables from this structure can proceed as follows: $\Pr(\pi_t^0 \mid o_t)$ is first used to sample $\pi_t^0$; $\mathcal{F}^{er}(s_t|\pi_t^0, o_t)$ is then used to sample $s_t$. Once we have obtained the sample for $\pi_t^0$ and $s_t$, we can proceed to sample the remaining nodes in the network $\mathcal{B}_t^{er}$ to obtain a sample for $l_t$ as usual. Finally, we note that the weight $w_t = \Pr(o_t)$ can also be computed efficiently by:

$$\Pr(o_t) = \sum_{\pi_t^0} \Pr(o_t \mid \pi_t^0) \Pr(\pi_t^0)$$

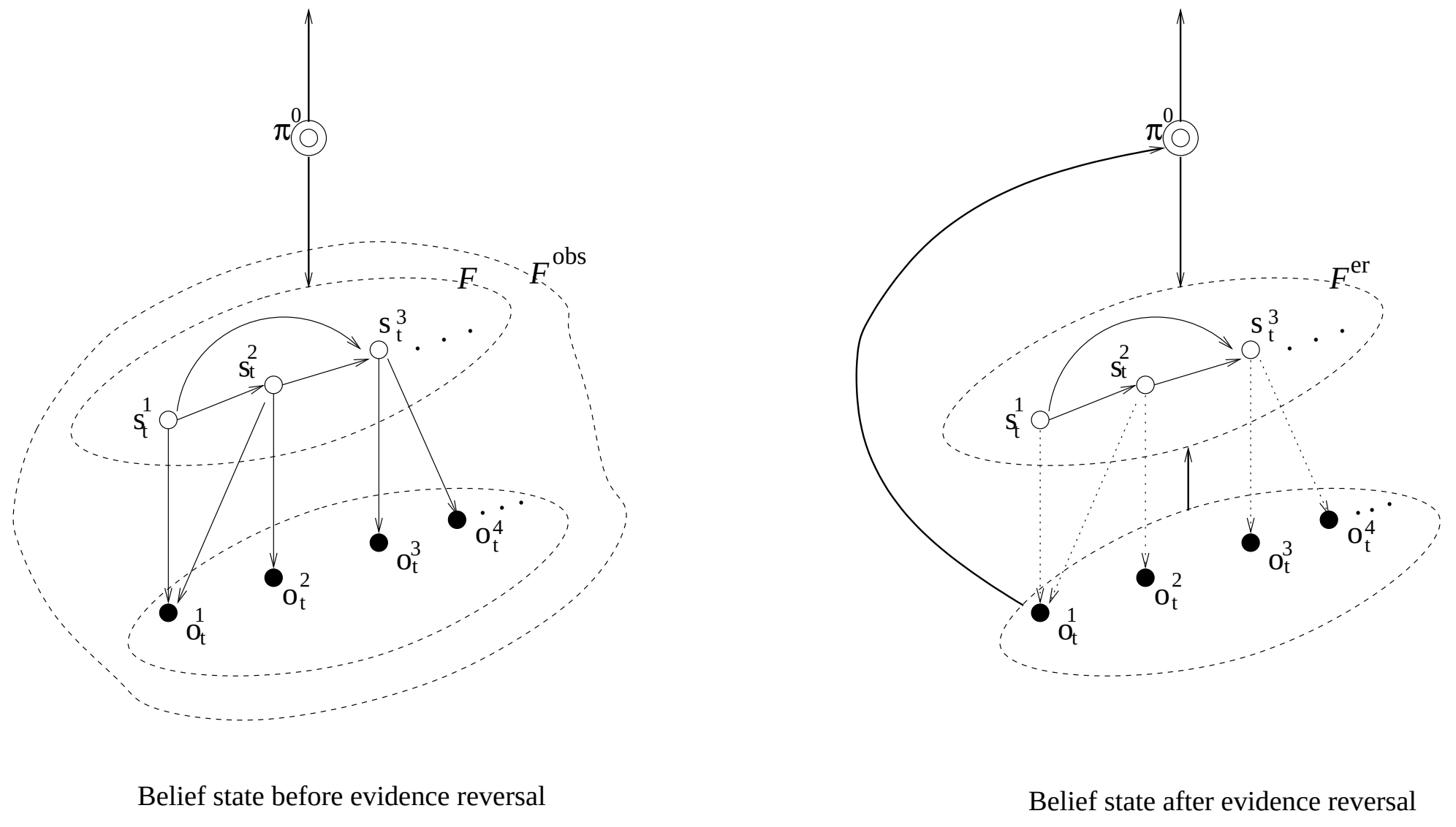

Figure 16: Evidence reversal with factored state space

In this evidence reversal procedure, for each value of $\pi_t^0$, we need to perform exact inference on the structure of $\mathcal{F}^{obs}(s_t, o_t|\pi_t^0)$. Thus the complexity of this procedure heavily depends on the complexity of the network structure of $\mathcal{F}$. However, as we have noted, due to the nature of the factored representation, $\mathcal{F}$ usually has a sparse structure so that exact inference can be performed efficiently. For example, in the special case where $\mathcal{F}$ is completely factored into the product of $M$ independent state variables which are then independently observed, the complexity becomes linear w.r.t. $M$.

## 6. Experimental Results

In this section, we present our experimental results with the policy recognition algorithm. In subsection 6.1, we demonstrate the effectiveness of the Rao-Blackwellised sampling method for policy recognition by comparing the performance of our Rao-Blackwellised procedure against likelihood weighting sampling in a synthetic tracking task. In subsection 6.2, we present an application of the AHMM framework to the problem of tracking human behaviours in a complex spatial environment using distributed video surveillance data.

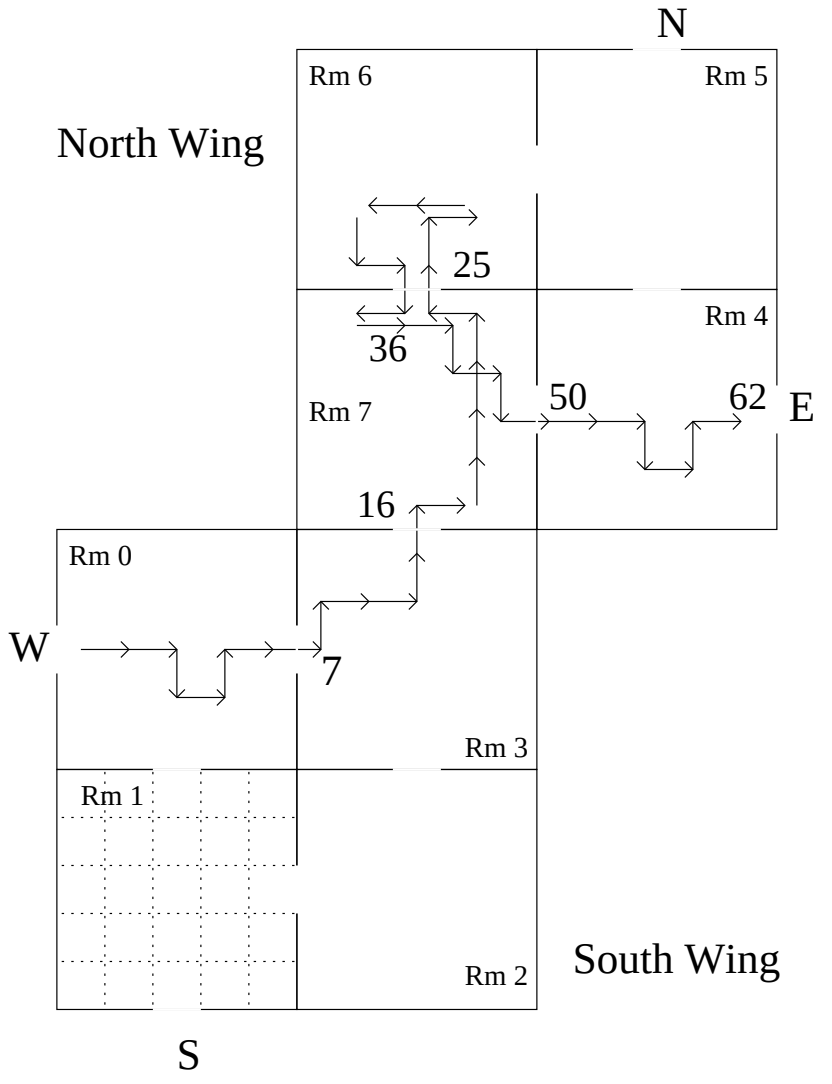

Figure 17: The environment and a sample trajectory

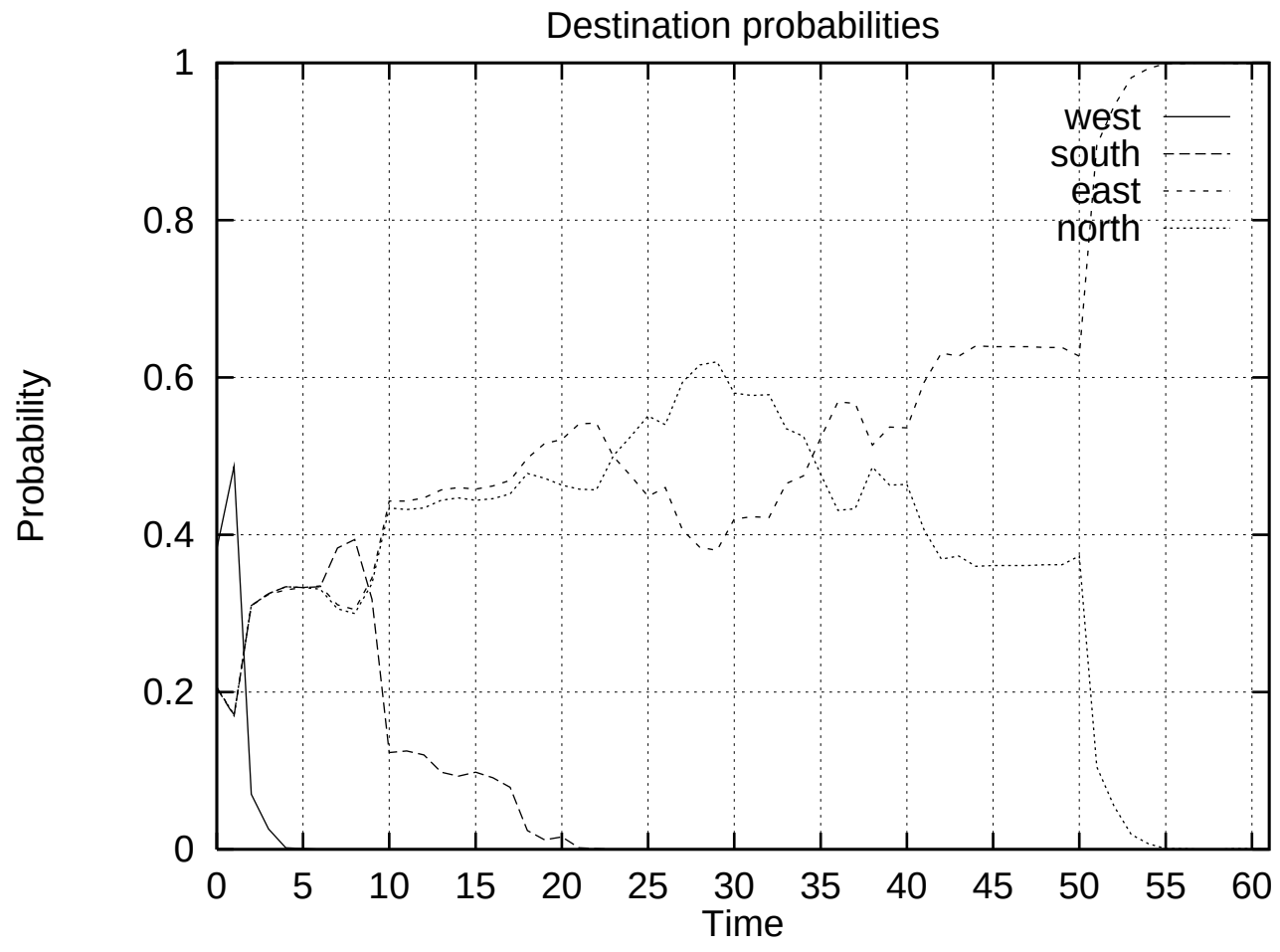

Figure 18: Probabilities of top-level destinations over time

### 6.1 Effectiveness of Rao-Blackwellisation

To demonstrate the effectiveness of the Rao-Blackwellised inference method for AHMM, we again consider the synthetic tracking task in which it is required to monitor and predict the movement of an agent through the building environment previously discussed in subsection 3.3. The structure of the AHMM used is the same as the one shown in Fig. 5. The parameters of the policies are chosen manually, and then used to simulate the movement of the agent in the building. To simulate the observation noise, we assume that the observation of the agent's true position can be anywhere among its 8 neighbouring cells with the probabilities given by a predefined observation model.

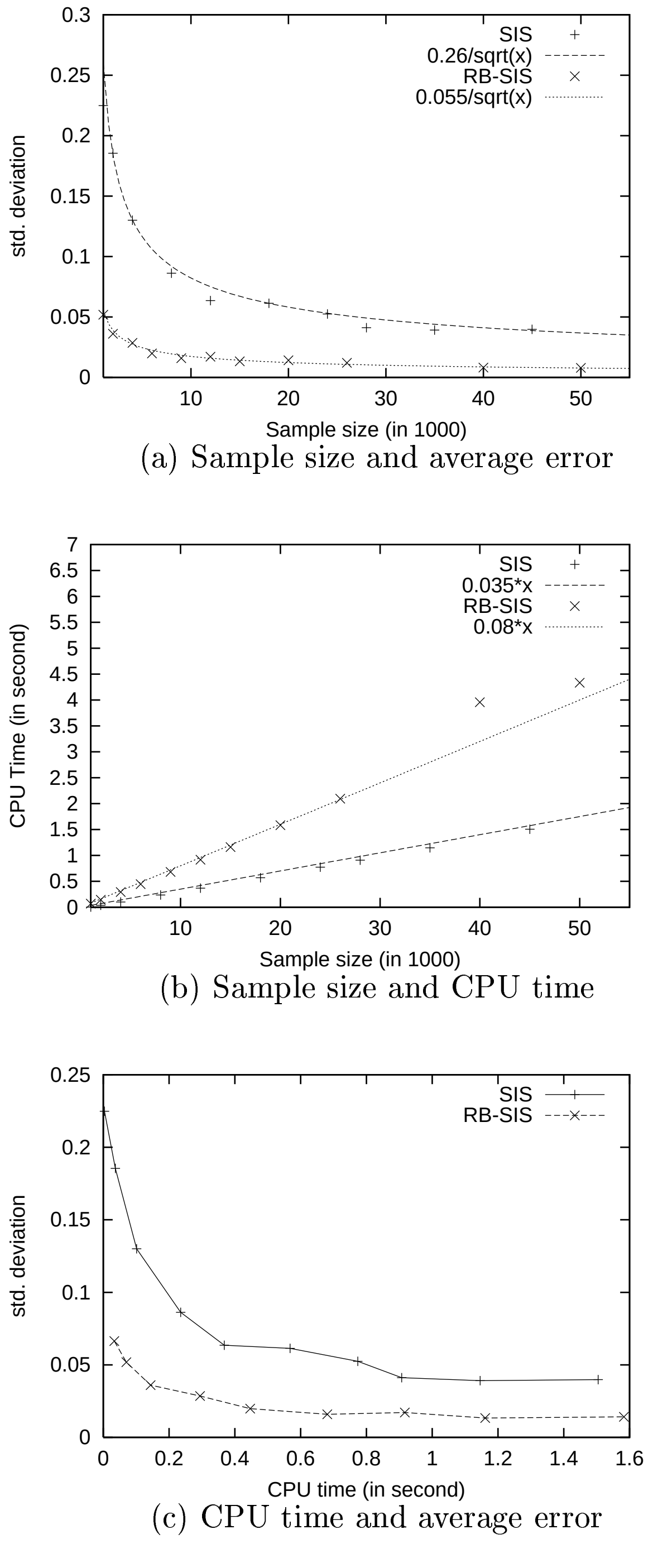

Figure 19: Performance profiles of SIS vs. RB-SIS

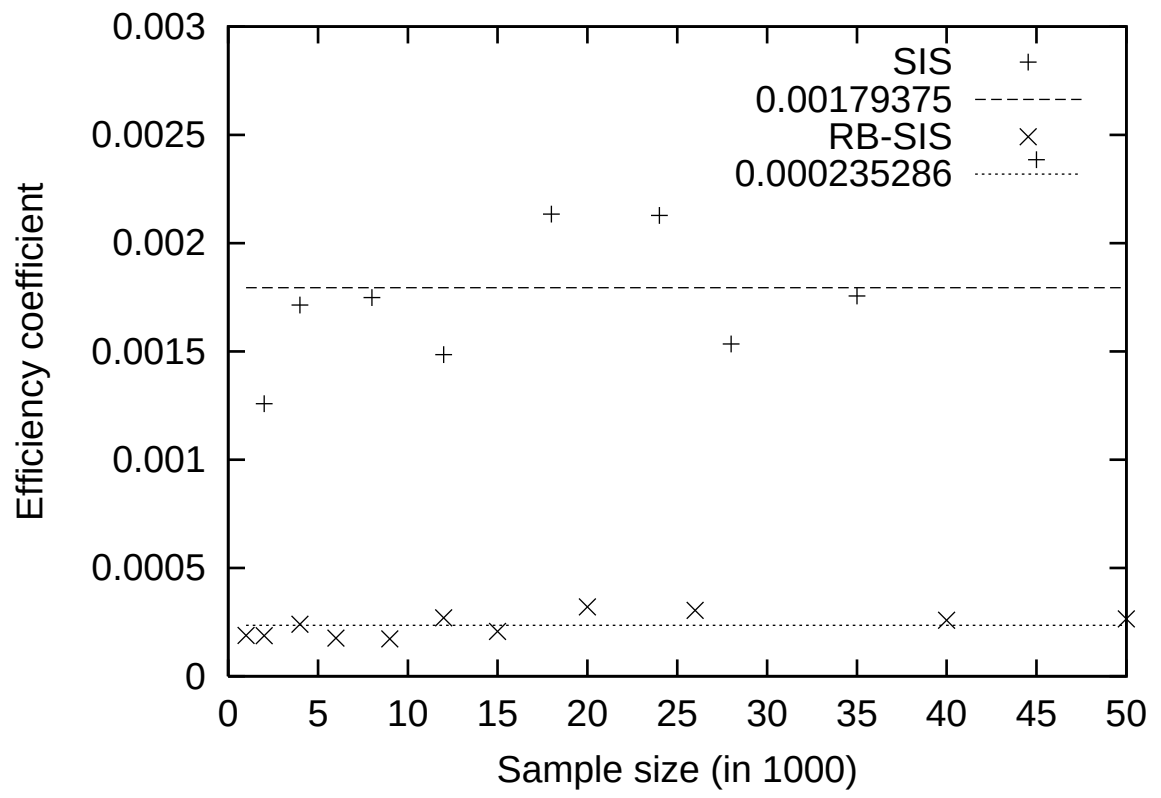

Figure 20: Efficiency coefficients of SIS and RB-SIS

We implement the RB-SIS method (with re-sampling) and use the policy hierarchy specification and the simulated observation sequence as input to the algorithm. In a typical run, the algorithm can return the probability of the main building exit, the next wing exit, and the next room-door that the agent is currently heading to. An example track is shown in Fig. 17. As the observations about the track arrive over time, the prediction probability distribution of which main building exit the track is heading to is shown in Fig. 18.

To illustrate the advantage of RB-SIS, we also implement an SIS method without Rao-Blackwellisation (LW with ER and re-sampling (Kanazawa et al., 1995)) and compare the performance of the two algorithms. We run the two algorithms using different sample population sizes to obtain their performance profiles. For a given sample size $N$, the standard deviation ($\sigma(N)$) over 50 runs in the estimated probabilities of the top-level policies is used as the measure of expected error in the probability estimates. We also record the average time taken in each update iteration.

Fig. 19(a) plots the standard deviation of the two algorithms for different sample sizes. The behaviour of the error follows closely the theoretical curve $\sigma(N) = c/\sqrt{N}$, or $\sigma^2(N) = c^2/N$, with $c_{SIS} \approx 0.26$ and $c_{RB-SIS} \approx 0.055$. As expected, for the same number of samples, the RB-SIS algorithm delivers much better accuracy.

Fig. 19(b) plots the average CPU time ($\mathcal{T}$) taken in each iteration versus the sample size. As expected, $\mathcal{T}(N)$ is linear to $N$, with the RB-SIS taking about twice longer due to the overhead in updating the RB belief state while processing each sample.

Fig. 19(c) plots the actual CPU time taken versus the expected error for the two algorithms. It shows that for the same CPU time spent, the RB-SIS method still significantly reduces the error in the probability estimates.

Note that for each algorithm, the quantity $\eta = \sigma^2(N)\mathcal{T}(N)$ is approximately constant since the dependency on $N$ cancels one another out. Thus, this constant can be used as an efficiency coefficient to measure the performance of the sampling algorithm independent of the number of samples. For example, if an algorithm has a twice smaller coefficient, it can deliver the same accuracy with half CPU time, or half the variance for the same CPU time. Fig. 20 plots the efficiency coefficients for both SIS and RB-SIS, with $\eta_{SIS} \approx 0.0018$ and $\eta_{RB-SIS} \approx 0.000235$. This indicates a performance gain of almost an order of magnitude (8 folds) for RB-SIS.

### 6.2 Application to Tracking Human Behaviours

Using the policy recognition algorithm, we have implemented a real-time surveillance system that tracks the behaviour of people in a complex indoor environment using surveillance video data. The environment consists of a corridor, the Vision lab and two offices (see Fig. 21). People enter/exit the scene via the left or the right entrance of the corridor. The system has six static cameras with overlapping field of views which cover most of the ground plane in the scene.

The entire environment is divided into a grid of cells, and the current cell position of the tracked object acts like the current state in our AHMM. The cameras are calibrated so that they can return the current position of the tracked object on the ground, however the returned coordinates are unreliable as the cameras have to deal with noisy video frames and occlusion of objects in the scene. For more information on how low-level tracking is done with multiple cameras, readers are referred to (Nguyen, Venkatesh, West, & Bui, 2002). We assume that the observation of a state can only be in the area surrounding it, thus the observation model is a matrix specifying the observation likelihood for each cell within a neighbourhood of the current state.

The policy hierarchy for behaviours in this environment is constructed as follows. First, we construct the region hierarchy with three levels. At the bottom level, we identify 7 regions of special interest: the corridor, the two offices, the areas surrounding the Linux server, NT server, printer, and the remaining free space in the Vision lab (Fig. 21). At the higher level, all regions in the Vision lab are grouped together. The top level consists of the entire environment. The policy hierarchy representing people's behaviors has three levels corresponding to the three levels of the region hierarchy (see Fig. 23). At the bottom level, we are interested in the behaviours that take place within each of the 7 regions of interest. For example, near the Linux server, the person might be using the Linux machine, or simply passing through that region, leading to two different policies. Similar policies are defined for the NT server region, the printer region, and the two small offices. In the corridor and inside the Vision lab (region 1 and 5), we construct different policies corresponding to the different destinations that the person is heading to. Region 5 also has a special policy representing the "walk-around" behaviour. At the middle level, three policies are defined for the corridor and office space representing a person's plan of exiting this space by the left/right entrance or by the door of the Vision lab. We define only one policy for the Vision lab to represent the typical behaviour of a lab user (e.g., go to Linux server, followed by go to printer).[8] Finally, for the top level region (the whole environment), we define two policies representing a person's leaving the scene via the left/right entrance.

Fig. 21 and 22 show two concurrent trajectories of two different people in this environment. Some sample video frames captured by the different cameras in the system are shown in Fig. 24.

With the AHMM model defined above, and a sequence of observations returned by the cameras, we first determine the performance profiles of RB-SIS and SIS in this real

8. If we consider different groups of lab users, each group might give rise to a different policy at this level.

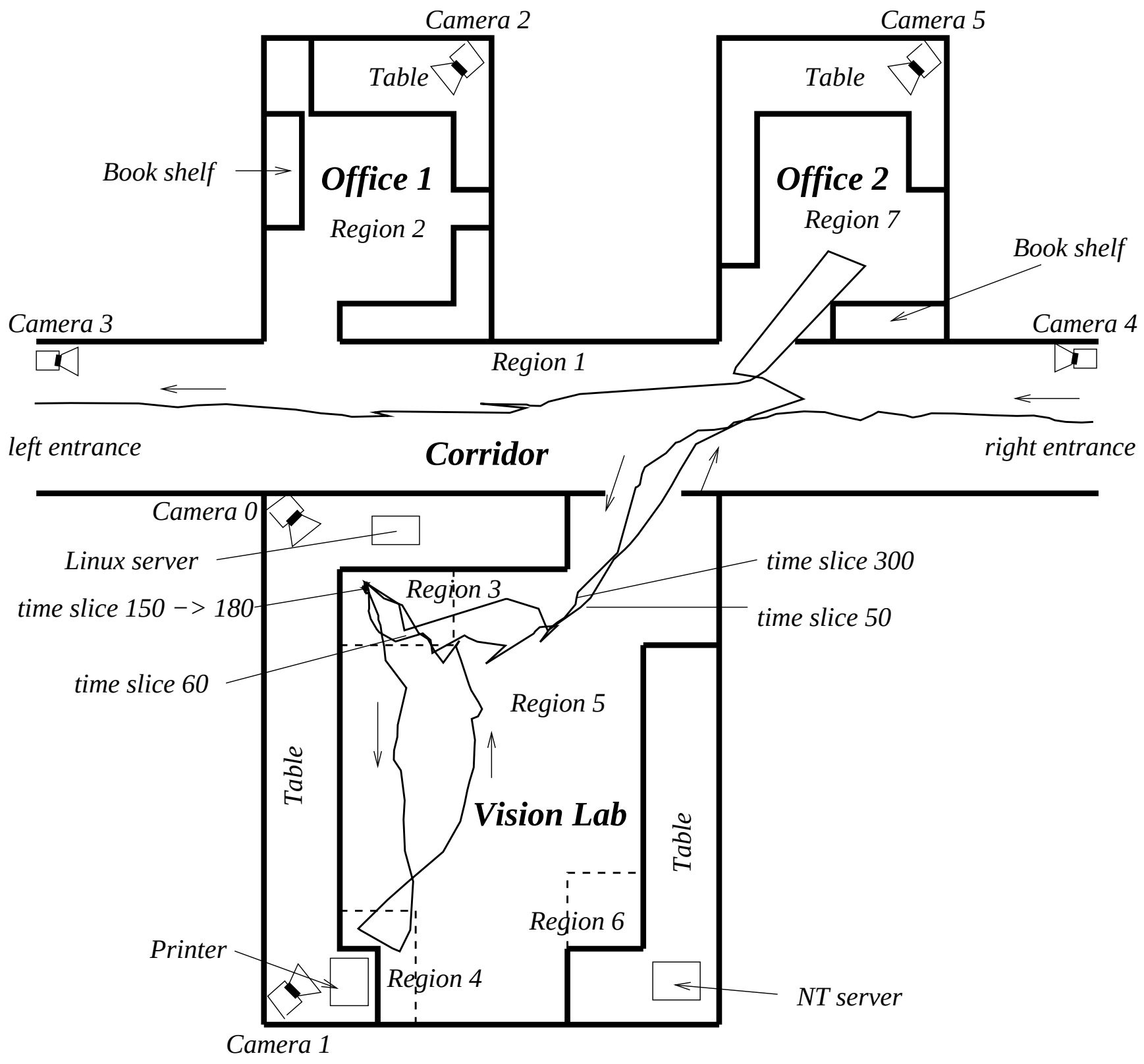

Figure 21: The environment and the trajectory of person 1

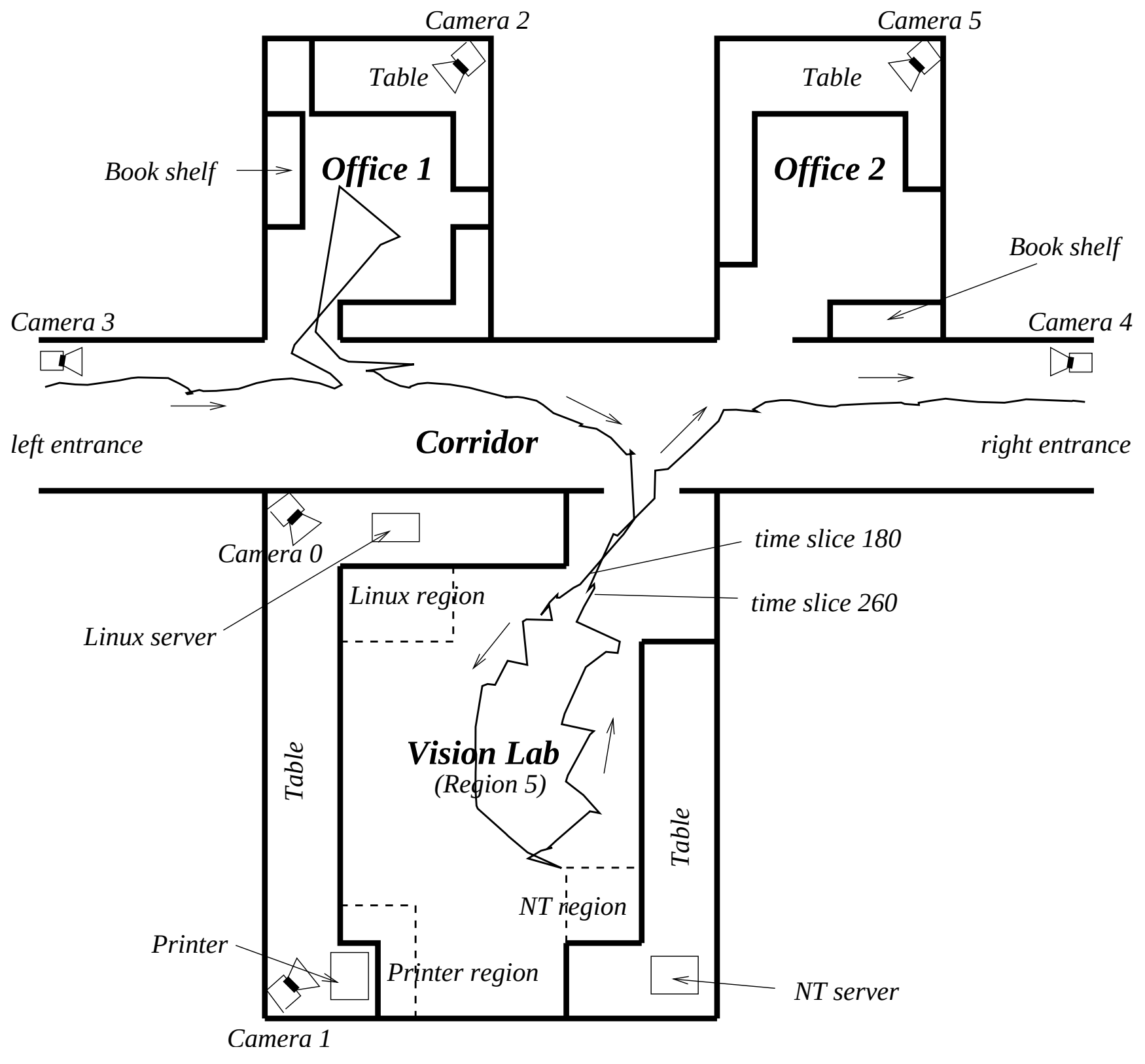

Figure 22: The trajectory of person 2

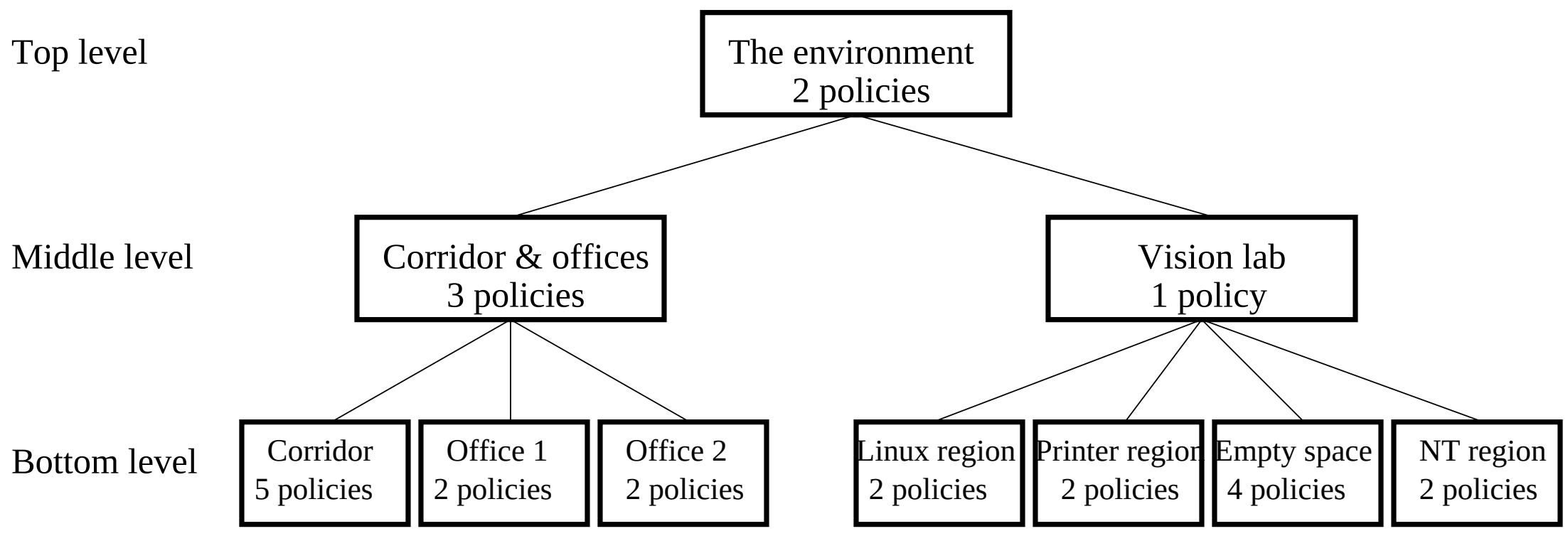

Figure 23: The region and policy hierarchy

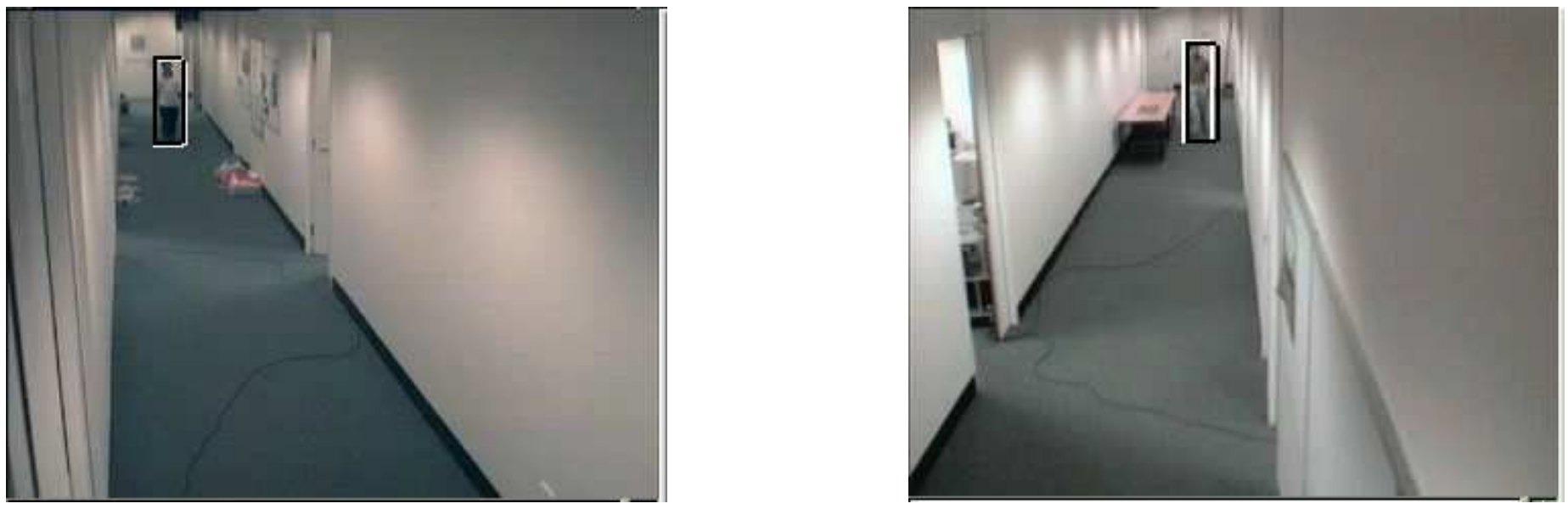

(a) (b)

Figure 24: (a) Person 1 enters the scene and (b) Person 2 enters the scene.

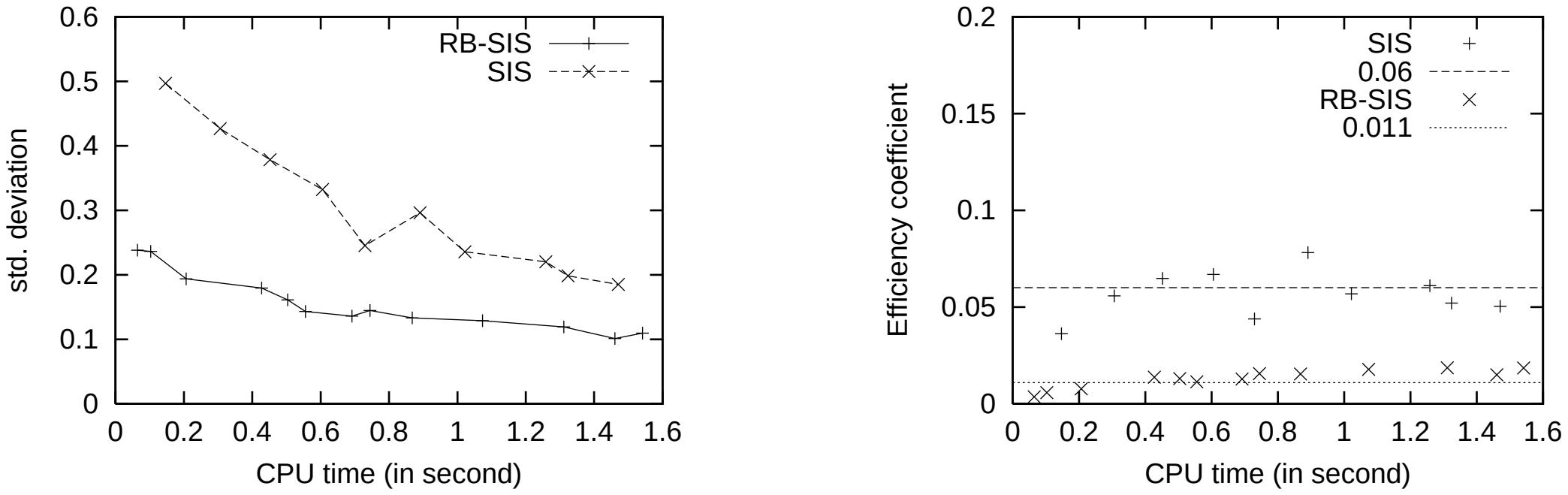

(a) Error vs. CPU time (b) Efficiency coefficients

Figure 25: Performance of RB-SIS and SIS with real tracking data

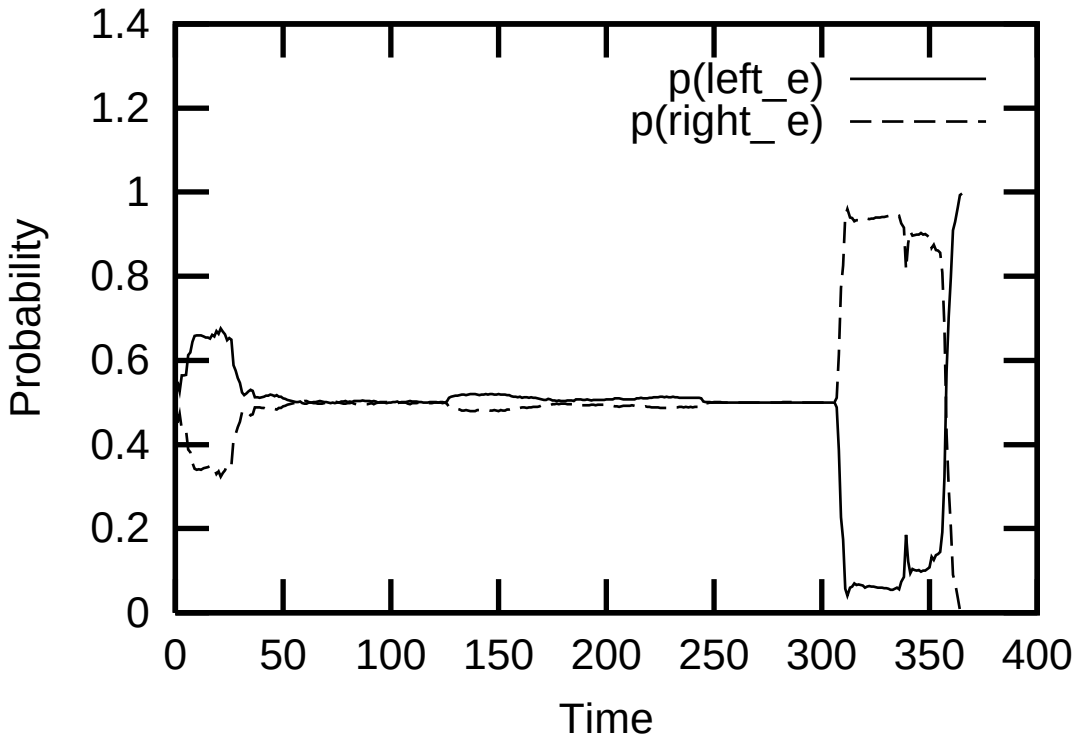

Figure 26: The probabilities that person 1 is leaving the scene via the entrances (top level policies)

environment. The two algorithms behave in a similar way as in the previous experiment with simulated data. Fig. 25 shows the error curve against the CPU time for the two algorithms. The efficiency co-efficient for RB-SIS in this case is $\eta_{RB-SIS} \approx 0.011$, and for SIS is $\eta_{SIS} \approx 0.06$. This shows that the RB-SIS still performs about 5 times better than SIS in this domain.

In the surveillance system, the low level tracking module returns the observations at the rate of approximately two per second. The observation is then passed to the RB-SIS algorithm which produces the probability estimate of the current policy at different levels in the hierarchy. At the moment, our surveillance system can run in real time using two AMD 1G machines. Examples of the output returned by the system for the two trajectories in Fig. 21 and Fig. 22 are given below.

Fig. 26 shows the probabilities that person 1 is exiting the environment by the left or right entrance (denoted by $p_{left_e}$ and $p_{right_e}$ respectively). At the beginning, $p_{left_e}$ increases when person 1 is heading to the left entrance (see the trajectory in Fig. 21). Then, $p_{left_e}$ is approximately constant from time slice 50 when person 1 is inside the Vision lab. This is because only one middle level policy is defined for the Vision lab and his movement inside the lab is independent of his final exit/entrance. At time slice 310, $p_{left_e}$ decreases when person 1 is leaving the lab, turning right, and entering office 2. Then, it increases and approaches 1 when he is leaving office 2, turning left, and going towards the left entrance. In contrast, $p_{right_e}$ falls quickly to zero during this time.

We now look at the results of querying of the bottom level policies. Fig. 27 shows the distribution of the possible destinations of person 2 from time slice 180 to time slice 260, when he is in region 5 (see the trajectory in Fig 22). The probabilities obtained show that the system is able to correctly detect the “walk-around” behaviour.

The final result (Fig. 28) shows the inferred behaviours of person 1 when he is at the Linux server region. Initially, the probabilities for “using Linux server” and for “passing through” are the same. As the person stays in the same position for an extended period of time, the system is able to identify the correct behaviour of person 1 as “using the Linux server”.

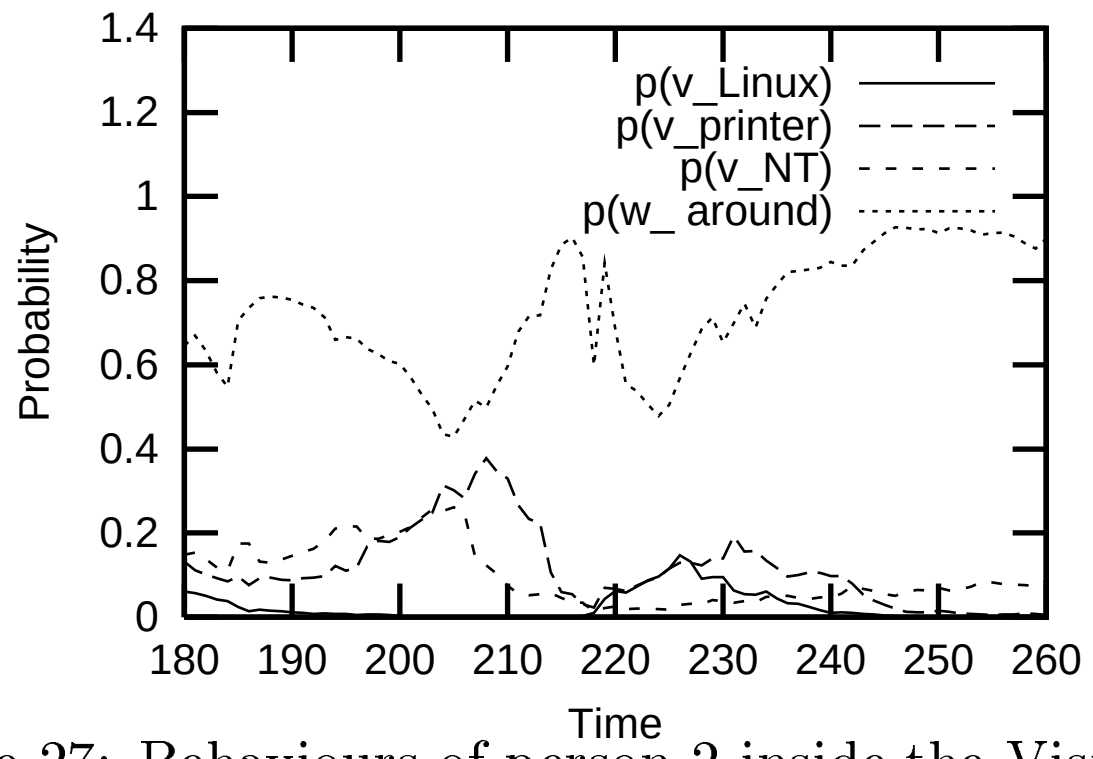

Figure 27: Behaviours of person 2 inside the Vision lab

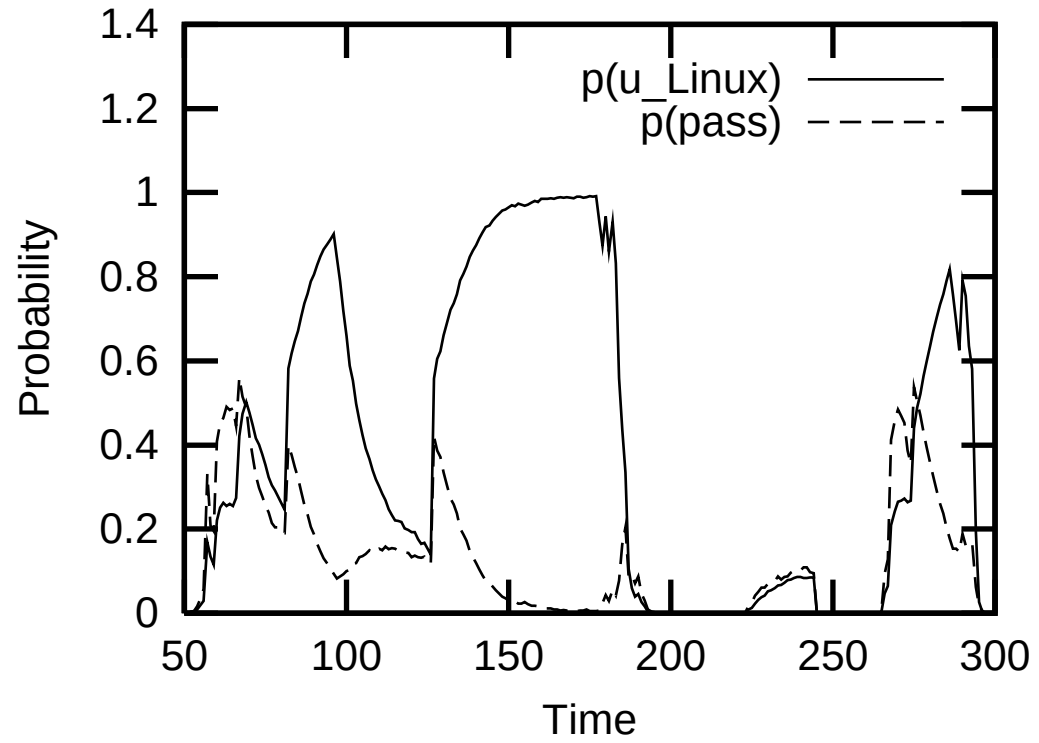

Figure 28: Behaviour of person 1 inside the Linux server region

## 7. Related Work in Probabilistic Plan Recognition

The case for using probabilistic inference for plan recognition has been argued convincingly by Charniak and Goldman (1993). However, the plan recognition Bayesian network used by Charniak and Goldman is a static network. Thus their approach would run into problems when they have to process on-line a stream of evidence about the plan. More recent approaches (Pynadath & Wellman, 1995, 2000; Goldman et al., 1999; Huber et al., 1994; Albrecht et al., 1998) have used dynamic stochastic models for plan recognition and thus are more suitable for doing on-line plan recognition under uncertainty.

Among these, the most closely related model to the AHMM is the Probabilistic State-Dependent Grammar (PSDG) (Pynadath, 1999; Pynadath & Wellman, 2000). A comparison of the representational aspect of the two models has been discussed under subsection 3.4. In terms of algorithms for plan recognition, Pynadath and Wellman only offer an exact method to deal with the case where the states are fully observable. When the states are partially observable, a brute-force approach is suggested which amounts to summing over all possible states. We note that even for the fully observable case, the belief state that we need to deal with can still be large since the policy starting/ending times are unknown.[9] Since an exact method is used by Pynadath and Wellman, the complexity for maintaining the belief state would most likely be exponential to the number of levels in the PSDG expansion hierarchy (i.e., the height of our policy hierarchy). On the other hand, our RB-SIS policy recognition algorithm can handle partially observable states and the Rao-Blackwellisation procedure ensures that the sampling algorithm scales well with the number of levels in the policy hierarchy. Furthermore, as we have noted in subsection 3.4, if we consider compound policies, the PSDG can be converted to an AHMM. In our framework, a compound policy $\pi^k = \pi^{k-1}_{(1)}, \ldots, \pi^{k-1}_{(m)}$ can be represented just as a normal policy, with a slight modification to let the variable $e^k$ take on values between 1 and $m+1$, where the value $m+1$ indicates that the compound policy has terminated. The policy recognition algorithm can then be modified to also work with this model.

Similar to our AHMM and the PSDG, the recent work by Goldman et al. (1999) also makes use of a detailed model of the plan execution process. Using the rich language of probabilistic Horn abduction, they are able to model more sophisticated plan structures such as interleaved/concurrent plans, partially-ordered plans. However the work serves mainly as a representational framework, and provides no analysis on the complexity of plan recognition in this setting.

Other work in probabilistic plan recognition up to date has employed much coarser models for plan execution. Most have ignored the important influence of the state of the world to the agent's planning decision (Goldman et al., 1999). To the best of our knowledge, none of the work up to date has addressed the problem of partial and noisy observation of the state. Most, except the PSDG, do not look at the observation of the outcomes of actions, and assume that the action can be observed directly and accurately. We note that this kind of simplifying assumptions is needed in previous work so that the computational complexity of performing probabilistic plan recognition remains manageable. In contrary, our work here illustrates that although the plan recognition dynamic stochastic model can

9. Of course, if an SRD policy hierarchy is considered then full observability alone is enough.

be complex, they exhibit special types of conditional independence which, if exploited, can lead to efficient plan recognition algorithms.

## 8. Conclusion and Future Work

In summary, we have presented an approach for on-line plan recognition under uncertainty using the AHMM as the model for the execution of a stochastic plan hierarchy and its noisy observation. The AHMM is a novel type of stochastic processes, capable of representing a rich class of plans and the associating uncertainty in the planning and plan observation process. We first analyse the AHMM structure and its conditional independence properties. This leads to the proposed hybrid Rao-Blackwellised Sequential Importance Sampling (RB-SIS) algorithm for performing belief state updating (filtering) for the AHMM which exploits the structure of the AHMM for greater efficiency and scalability. We show that the complexity of RB-SIS when applied to the AHMM only depends linearly on the number of levels $K$ in the policy hierarchy, while the sampling error does not depend on $K$.

In terms of plan recognition, these results show that while the stochastic process for representing the execution of a plan hierarchy can be complex, they exhibit certain conditional independence properties that are inherent in the dynamics of the planning and acting process. These independence properties, if exploited, can help to reduce the complexity of performing inference on the plan execution stochastic model, leading to feasible and scalable algorithms for on-line plan recognition in noisy and uncertain domains. The scalability of the algorithm for policy recognition provides the possibility to consider more complex plan hierarchies and more detailed models of the plan execution process. The key to achieve this efficiency, as we have shown in the paper, is a combination of recently developed techniques in probabilistic inference: compact representations for Bayesian networks (context-sensitive independence, factored representations), and hybrid DBN inference which can take advantage of these compact representations (Rao-Blackwellisation).

Several future research directions are possible. To further investigate the AHMM, we would like to consider the problem of learning the parameters of an AHMM from a database of observation sequences, e.g., to learn the plan execution model by observing multiple episodes of an agent executing the same plan. The structure of the AHMM suggests that we can try to learn the model of each abstract policy separately. Indeed, if we can observe the execution of each abstract policy separately, the learning problem is reduced to HMM parameter re-estimation for level-1 policies, and simple frequency counting for higher-level policies. If the observation sequence is a long episode with no clear cut temporal boundary between the policies, the problem becomes a type of parameter estimation for DBN with hidden variables, and techniques for dealing with hidden variables such as EM (Dempster, Laird, & Rubin, 1977) can be applied.

Extensions can be made to the AHMM to make the model more expressive and suitable for representing more complex agents' plans. For example, a more expressive plan execution model such as the HAM model (Parr, 1998) can be considered so that state-independent sequences of policies can be represented. The current model can also be enriched to consider a set of top-level policies which can be interleaved during their execution. We expect that these new models would exhibit context-specific independence properties similar to the AHMM, and Rao-Blackwellised sampling methods for policy recognition in these models can be derived.

## Acknowledgement

We would like to thank the anonymous reviewers for their insightful comments which have helped improve both the presentation and the contents of this paper. Many thanks to Nam Nguyen for his implementation of the distributed tracking system used in this paper.

## References

Albrecht, D. W., Zukerman, I., & Nicholson, A. E. (1998). Bayesian models for keyhole plan recognition in an adventure game. *User Modelling and User-adapted Interaction*, *8*(1–2), 5–47.

Andrieu, C., & Doucet, A. (2000). Particle filtering for partially observed Gaussian state space models. Tech. rep. CUED-F-INFENG/TR. 393, Signal Processing Group, University of Cambridge, Cambridge, UK.

Bauer, M. (1994). Integrating probabilistic reasoning into plan recognition. In *Proceedings of the Eleventh European Conference on Artificial Intelligence.*

Boutilier, C., Dearden, R., & Goldszmidt, M. (2000). Stochastic dynamic programming with factored representations. *Artificial Intelligence*, *121*, 49–107.

Boutilier, C., Friedman, N., Goldszmidt, M., & Koller, D. (1996). Context-specific independence in Bayesian networks. In *Proceedings of the Twelveth Annual Conference on Uncertainty in Artificial Intelligence.*

Boyen, X., & Koller, D. (1998). Tractable inference for complex stochastic processes. In *Proceedings of the Fourteenth Annual Conference on Uncertainty in Artificial Intelligence.*

Brand, M. (1997). Coupled hidden Markov models for modeling interacting processes. Tech. rep. 405, MIT Media Lab.

Bui, H. H., Venkatesh, S., & West, G. (1999). Layered dynamic Bayesian networks for spatio-temporal modelling. *Intelligent Data Analysis*, *3*(5), 339–361.

Bui, H. H., Venkatesh, S., & West, G. (2000). On the recognition of abstract Markov policies. In *Proceedings of the National Conference on Artificial Intelligence (AAAI-2000)*, pp. 524–530.

Casella, G., & Robert, C. P. (1996). Rao-Blackwellisation of sampling schemes. *Biometrika*, *83*, 81–94.

Castillo, E., Gutierrez, J. M., & Hadi, A. S. (1997). *Expert systems and probabilistic network models.* Springer.

Charniak, E., & Goldman, R. (1993). A Bayesian model of plan recognition. *Artificial Intelligence*, *64*, 53–79.

Cohen, P. R., & Levesque, H. J. (1990). Intention is choice with commitment. *Artificial Intelligence*, *42*, 213–261.

Cooper, G. F. (1990). The computational complexity of probabilistic inference using Baysian belief networks. *Artificial Intelligence*, *42*, 393–405.

Dagum, P., & Luby, M. (1993). Approximating probabilistic inference in Bayesian belief networks is NP-hard. *Artificial Intelligence*, *60*, 141–153.

Dagum, P., Galper, A., & Horvitz, E. (1992). Dynamic network models for forecasting. In *Proceedings of the Eighth Annual Conference on Uncertainty in Artificial Intelligence*, pp. 41–48.

D'Ambrosio, B. (1993). Incremental probabilistic inference. In *Proceedings of the Ninth Annual Conference on Uncertainty in Artificial Intelligence*, pp. 301–308.

Dawid, A. P., Kjærulff, U., & Lauritzen, S. (1995). Hybrid propagation in junction trees. In Zadeh, L. A. (Ed.), *Advances in Intelligent Computing*, Lecture Notes in Computer Science, pp. 87–97.

Dean, T., & Kanazawa, K. (1989). A model for reasoning about persistence and causation. *Computational Intelligence*, *5*(3), 142–150.

Dean, T., & Lin, S.-H. (1995). Decomposition techniques for planning in stochastic domains. In *Proceedings of the Fourteenth International Joint Conference on Artificial Intelligence (IJCAI-95)*.

Dempster, A., Laird, N., & Rubin, D. (1977). Maximum likelihood from incomplete data via the EM algorithm. *Journal of the Royal Statistical Society B*, *39*, 1–38.

Doucet, A., de Freitas, N., Murphy, K., & Russell, S. (2000a). Rao-Blackwellised particle filtering for dynamic Bayesian networks. In *Proceedings of the Sixteenth Annual Conference on Uncertainty in Artificial Intelligence.*

Doucet, A., Godsill, S., & Andrieu, C. (2000b). On sequential Monte Carlo sampling methods for Bayesian filtering. *Statistics and Computing*, *10*(3), 197–208.

Forestier, J.-P., & Varaiya, P. (1978). Multilayer control of large Markov chains. *IEEE Transactions on Automatic Control*, *23*(2), 298–305.

Fung, R., & Chang, K. C. (1989). Weighting and integrating evidence for stochastic simulation in bayesian networks. In *Proceedings of the Fifth Conference on Uncertainty in Artificial Intelligence.*

Geweke, J. (1989). Bayesian inference in econometric models using Monte Carlo integration. *Econometrica*, *57*(6), 1317–1339.

Ghahramani, Z., & Jordan, M. I. (1997). Factorial hidden Markov models. *Machine Learning*, *29*, 245–273.

Goldman, R., Geib, C., & Miller, C. (1999). A new model of plan recognition. In *Proceedings of the Fifteenth Annual Conference on Uncertainty in Artificial Intelligence.*

Hauskrecht, M., Meuleau, N., Kaelbling, L. P., Dean, T., & Boutilier, C. (1998). Hierarchical solution of Markov decision processes using macro-actions. In *Proceedings of the Fourteenth Annual Conference on Uncertainty in Artificial Intelligence.*

Henrion, M. (1988). Propagating uncertainty in Bayesian networks by probabilistic logic sampling. In Lemmer, J., & Kanal, L. (Eds.), *Uncertainty in Artificial Intelligence 2*, Amsterdam. North-Holland.

Huber, M. J., Durfee, E. H., & Wellman, M. P. (1994). The automated mapping of plans for plan recognition. In *Proceedings of the Tenth Annual Conference on Uncertainty in Artificial Intelligence*.

Jelinek, F., Lafferty, J. D., & Mercer, R. L. (1992). Basic methods of probabilistic context free grammar. In Laface, P., & Mori, R. D. (Eds.), *Recent Advances in Speech Recognition and Understanding*, pp. 345–360. Springer-Verlag.

Jensen, F. (1996). *An Introduction to Bayesian Networks*. Springer.

Jensen, F., Lauritzen, S., & Olesen, K. (1990). Bayesian updating in recursive graphical models by local computations. *Computational Statistics Quarterly*, *4*, 269–282.

Jordan, M. I., Ghahramani, Z., Jaakkola, T. S., & Saul, L. K. (1999). An introduction to variational methods for graphical models. *Machine learning*, *37*(2), 183–233.

Jordan, M. I., Ghahramani, Z., & Saul, L. K. (1997). Hidden Markov decision trees. In Mozer, M. C., Jordan, M. I., & Petsche, T. (Eds.), *Advances in Neural Information Processing Systems 9*, Cambridge, MA. MIT Press.

Kalman, R. E. (1960). A new approach to linear filtering and prediction problems. *Transactions of the American Society of Mechanical Engineering, Series D, Journal of Basic Engineering*, *82*, 35–45.

Kanazawa, K., Koller, D., & Russell, S. (1995). Stochastic simulation algorithms for dynamic probabilistic networks. In *Proceedings of the Eleventh Annual Conference on Uncertainty in Artificial Intelligence*, pp. 346–351.

Kautz, H., & Allen, J. F. (1986). Generalized plan recognition. In *Proceedings of the Fifth National Conference on Artificial Intelligence*, pp. 32–38.

Kjaerulff, U. (1992). A computational scheme for reasoning in dynamic probabilistic networks. In *Proceedings of the Eighth Annual Conference on Uncertainty in Artificial Intelligence*, pp. 121–129.

Kjærulff, U. (1995). dHugin: A computational system for dynamic time-sliced Bayesian networks. *International Journal of Forecasting*, *11*, 89–111.

Lauritzen, S., & Spiegelhalter, D. (1988). Local computations with probabilities on graphical structures and their application to expert systems. *Journal of the Royal Statistical Society B*, *50*, 157–224.

Liu, J. S., & Chen, R. (1998). Sequential Monte Carlo methods for dynamic systems. *Journal of the American Statistical Association*, *93*, 1032–1044.

Murphy, K., & Russell, S. (2001). Rao-blackwellised particle filtering for dynamic Bayesian networks. In Doucet, A., de Freitas, N., & Gordon, N. J. (Eds.), *Sequential Monte Carlo Methods in Practice*. Springer-Verlag.

Murphy, K. P. (2000). Bayesian map learning in dynamic environments. In *Advances in Neural Information Processing Systems 12*, pp. 1015–1021. MIT Press.

Nguyen, N. T., Venkatesh, S., West, G., & Bui, H. H. (2002). Coordination of multiple cameras to track multiple people. In *Proceedings of the Asian Conference on Computer Vision (ACCV-2002)*, pp. 302–307.

Nicholson, A. E., & Brady, J. M. (1992). The data association problem when monitoring robot vehicles using dynamic belief networks. In *Proceedings of the Tenth European Conference on Artificial Intelligence*, pp. 689–693.

Parr, R. (1998). *Hierarchical control and learning for Markov Decision Processes.* Ph.D. thesis, University of California, Berkeley.

Parr, R., & Russell, S. (1997). Reinforcement learning with hierarchies of machines. In *Advances in Neural Information Processing Sytems (NIPS-97).*

Pearl, J. (1988). *Probabilistic Reasoning in Intelligent Systems: Networks of Plausible Inference.* Morgan Kaufmann, San Mateo, CA.

Pearl, J. (1987). Evidential reasoning using stochastic simulation of causal models. *Artificial Intelligence*, *32*, 245–257.

Pynadath, D. V. (1999). *Probabilistic grammars for plan recognition.* Ph.D. thesis, Computer Science and Engineering, University of Michigan.

Pynadath, D. V., & Wellman, M. P. (1995). Accounting for context in plan recognition, with application to traffic monitoring. In *Proceedings of the Eleventh Annual Conference on Uncertainty in Artificial Intelligence.*

Pynadath, D. V., & Wellman, M. P. (2000). Probabilistic state-dependent grammars for plan recognition. In *Proceedings of the Sixteenth Annual Conference on Uncertainty in Artificial Intelligence.*

Rabiner, L. R. (1989). A tutorial on Hidden Markov Models and selected applications in speech recognition. *Proceedings of the IEEE*, *77*(2), 257–286.

Sacerdoti, E. (1974). Planning in a hierarchy of abstraction spaces. *Artificial Intelligence*, *5*, 115–135.

Shachter, R. (1986). Evaluating influence diagrams. *Operations Research*, *34*, 871–882.

Shachter, R. D., & Peot, M. A. (1989). Simulation approaches to general probabilistic inference on belief networks. In *Proceedings of the Fifth Conference on Uncertainty in Artificial Intelligence.*

Sutton, R. S. (1995). Td models: Modelling the world at a mixture of time scales. In *Proceedings of the Internation Conference on Machine Learning (ICML-95).*

Sutton, R. S., Precup, D., & Singh, S. (1999). Between MDP and semi-MDPs: A framework for temporal abstraction in reinforcement learning. *Artificial Intelligence*, *112*, 181–211.

van Beek, P. (1996). An investigation of probabilistic interpretations of heuristics in plan recognition. In *Proceedings of the Fifth International Conference on User Modeling*, pp. 113–120.

York, J. (1992). Use of Gibbs sampler in expert systems. *Artificial Intelligence*, *56*, 115–130.